%% file: main.tex
\documentclass[11pt]{article}

\usepackage[final]{acl}

\usepackage{times}
\usepackage{latexsym}

\usepackage[T1]{fontenc}

\usepackage[utf8]{inputenc}

\usepackage{microtype}

\usepackage{inconsolata}

\usepackage{hyperref}
\usepackage{url}
\usepackage{enumitem}
\usepackage{booktabs} 
\usepackage{multirow}
\usepackage{graphicx}
\usepackage{enumitem} 
\usepackage[table]{xcolor}
\definecolor{mygray}{gray}{0.93} 
\usepackage{graphicx}    
\usepackage{subcaption}  
\usepackage{wrapfig}
\usepackage{algorithm}
\usepackage{algorithmic}
\usepackage{amsmath}
\usepackage{amssymb}
\usepackage{colortbl}
\usepackage{xcolor}

\usepackage{enumitem}  
\newcommand{\ie}{\emph{i.e., }}
\newcommand{\eg}{\emph{e.g., }}

\title{Answer Probing-Guided Search for Diverse Solution Exploration of LLMs}

\author{
Yi Fang\textsuperscript{1,2}, \enspace
Que Shen\textsuperscript{3}, \enspace
Chengpeng Li\textsuperscript{3}, \enspace
Boyi Deng\textsuperscript{1}, \enspace
Wei Shi\textsuperscript{4}, \\
\bfseries
Wenjie Wang\textsuperscript{1}\thanks{~Corresponding authors.}, \enspace
Fuli Feng\textsuperscript{1}, \enspace
Fengli Xu\textsuperscript{2,5}\footnotemark[1], \enspace
Dayiheng Liu\textsuperscript{3} \\
\\
\textsuperscript{1}University of Science and Technology of China \quad
\textsuperscript{2}Zhongguancun Academy \\
\textsuperscript{3}Alibaba Group \quad
\textsuperscript{4}Shanghai Jiao Tong University \quad
\textsuperscript{5}Tsinghua University \\
}

\begin{document}
\maketitle

\input{chapters/0_abstract}

\input{chapters/1_intro}

\input{chapters/2_preliminary}

\input{chapters/3_method}

\input{chapters/4_experiments}
\input{chapters/5_related_work}

\input{chapters/6_conclusion}

\section*{Acknowledgments}
This work is supported by the Zhongguancun Academy (Grant No. C20250401).

\bibliography{main}

\appendix
\input{chapters/7_appendix}

\end{document}

%% file: chapters/0_abstract.tex
\begin{abstract}

Generating multiple diverse and high-quality solutions is valuable for many applications, such as code-test generation and drug discovery. However, Large Language Models (LLMs) tend to converge on a single high-confidence solution during inference, limiting exploration of alternative valid solution paths.
Existing test-time methods promote diversity through tree-like search and prune semantically similar branches using response-level semantic embeddings. However, we find that such embeddings are easily confounded by linguistic and stylistic similarities, making it difficult to distinguish genuinely distinct solution paths.
To address this, we introduce \textbf{Answer Probing}, which probes the potential answer an LLM would reach from an intermediate reasoning path. 
We demonstrate that the hidden states of probed answers more effectively differentiate distinct solution paths than semantic embeddings, and the perplexity of probed answers serves as a practical proxy for reasoning correctness.
Based on these findings, we propose \textbf{A}nswer \textbf{P}robing-Guided \textbf{T}ree \textbf{S}earch (APTS), which guides the tree search by the probed answers’ hidden state similarity and perplexity.
Experiments on three reasoning tasks across two LLMs show that APTS consistently enhances solution diversity, demonstrating its effectiveness and robustness.

\end{abstract}

%% file: chapters/1_intro.tex
\section{Introduction}

Diverse solution generation requires Large Language Models (LLMs) to explore various solution paths and produce multiple high-quality solution candidates. This capability is highly valuable in real-world tasks where a set of high-quality candidates is needed rather than a single gold standard answer. For example, in code-test generation~\citep{JainSR25}, diverse test cases enable more comprehensive software evaluation. In drug discovery~\citep{drug_discover}, diverse candidates provide a broader range of chemical structures, helping experts identify novel drug leads. However, during inference, current LLMs tend to converge on a single high-confidence trajectory when sampling multiple times~\citep{RL_Sim, jiang2025artificial}, limiting their ability to explore other valid solutions. To address this issue, it is important to develop methods that encourage LLMs to explore more diverse yet valid reasoning paths.

\begin{figure}[t]
    \centering
    \includegraphics[width=\linewidth]{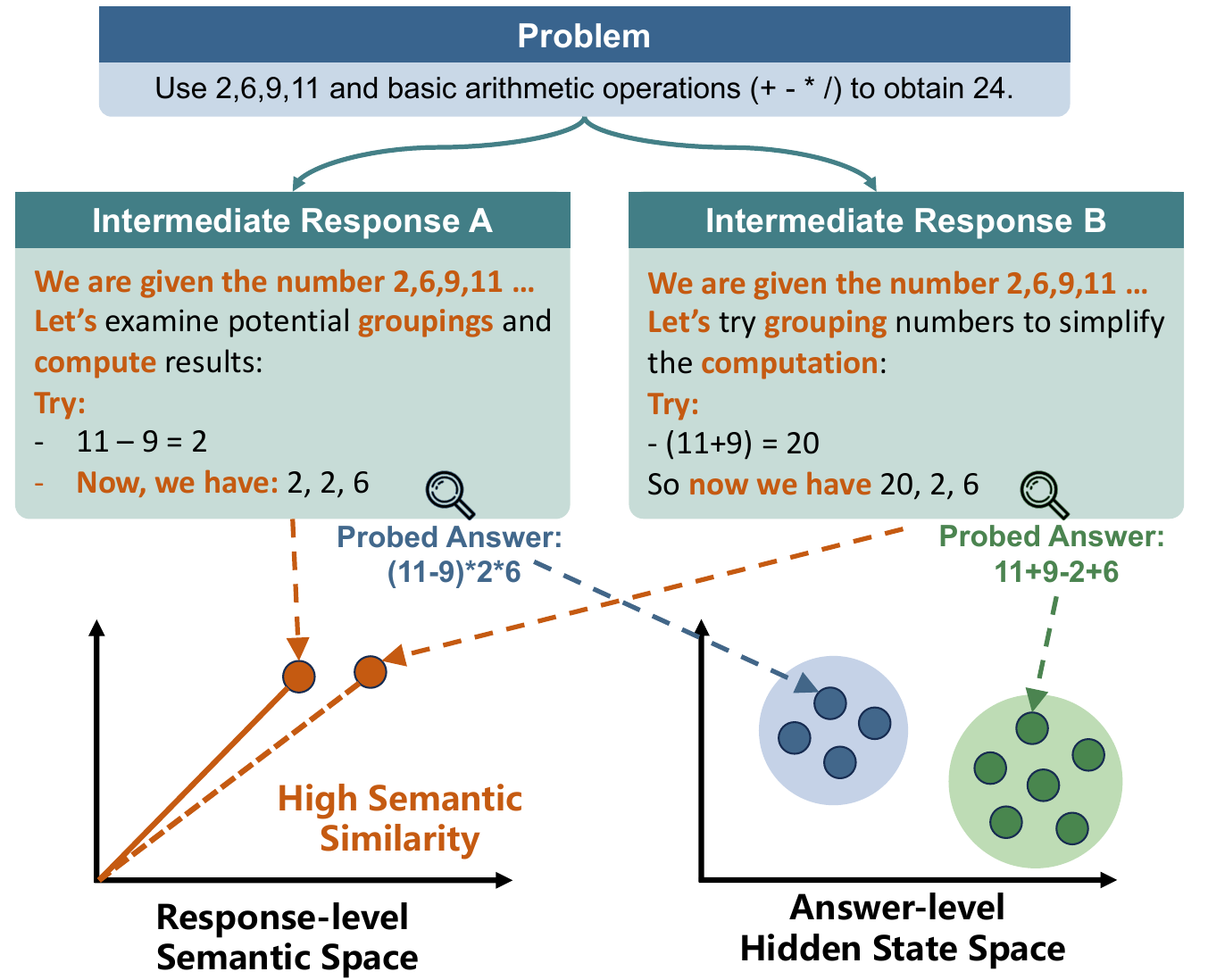}
    \caption{Illustration of Answer Probing. Compared with semantic embeddings, answer probing mitigates linguistic and stylistic noise in reasoning and better distinguishes different solution paths.}
    \label{fig:teaser}
\end{figure}

To improve reasoning diversity without incurring prohibitive training costs, several test-time methods have been explored. These approaches can be broadly categorized into decoding-based and search-based methods. Decoding-based methods, such as temperature sampling~\citep{renze2024effect} and nucleus sampling~\citep{HoltzmanBDFC20}, enhance diversity by introducing probabilistic perturbations into the token generation process. However, these methods often yield token-level variation without true semantic diversity: the generated reasoning paths may exhibit different token sequences but still reflect similar lines of reasoning~\citep{YunAWPS25}. Search-based methods address this limitation by leveraging semantic embeddings to guide the generation process~\citep{SemDiD, DiverseBeam}. Specifically, they employ search algorithms such as beam search~\citep{beam_search} and Tree of Thoughts~\citep{ToT} to explore multiple candidate paths at each reasoning step. These paths are then pruned based on their semantic embedding similarity, retaining only semantically diverse branches for further reasoning, thereby improving the diversity of the final reasoning set. However, LLM responses to the same problem frequently exhibit high structural and stylistic homogeneity~\citep{kirk2023understanding, o2024attributing}. 
As illustrated in Figure~\ref{fig:teaser}, shared reasoning signposts and surface patterns can induce high similarity in semantic space, reducing the effectiveness of semantic embeddings in identifying truly distinct solution paths (see detailed analysis in Section~\ref{sec:preliminary}).


To address this limitation, we seek an alternative representation that better reflects underlying solution paths while being less affected by linguistic and stylistic noise. Our preliminary analysis (Section~\ref{sec:preliminary}) shows that answer-level hidden states are more effective than response-level semantic embeddings at distinguishing divergent solution paths.
Motivated by this finding, we propose \textbf{Answer Probing}: at each reasoning step, we probe the potential answers the LLM would reach from the current intermediate reasoning path. 
This process exposes the latent solution tendency encoded in the intermediate reasoning path. 
By representing candidate reasoning paths with the mean-pooled hidden states of probed answers, we can better distinguish paths leading to distinct solutions. 
We also find that the perplexity (PPL) of probed answers provides a useful proxy for the quality of a candidate reasoning path, with lower PPL generally associated with more accurate outcomes.

Based on these findings, we propose \textbf{A}nswer \textbf{P}robing-Guided \textbf{T}ree \textbf{S}earch (APTS), which improves solution diversity while maintaining solution quality during generation. At each tree search step, APTS performs answer probing to evaluate candidate nodes. It uses the PPL of the probed answer as a quality signal and the hidden state similarity between probed answers as a diversity signal. By selecting nodes that are both high quality and non-redundant for further expansion, APTS guides the search toward candidate paths that are both promising and diverse. Extensive evaluations on three cross-domain reasoning tasks and two LLMs show that APTS improves solution diversity with only a modest impact on accuracy, demonstrating its effectiveness and robustness. In summary, the contributions of this work are threefold:
\begin{itemize}[itemsep=1pt, topsep=2pt, parsep=0pt, leftmargin=*]
    \item We propose \textbf{Answer Probing}, which probes intermediate answers to yield both a quality signal and a diversity signal, defining diversity at the level of underlying solution tendency rather than text-level semantic similarity.
    \item We propose \textbf{APTS}, a tree search algorithm which guides search using probed-answer hidden state similarity and PPL to improve solution diversity.
    \item We conduct extensive experiments across three domains and two LLM architectures, showing that APTS consistently improves diversity with minimal impact on accuracy.
\end{itemize}

%% file: chapters/2_preliminary.tex
\begin{figure*}[t]
    \centering
    \includegraphics[width=\linewidth]{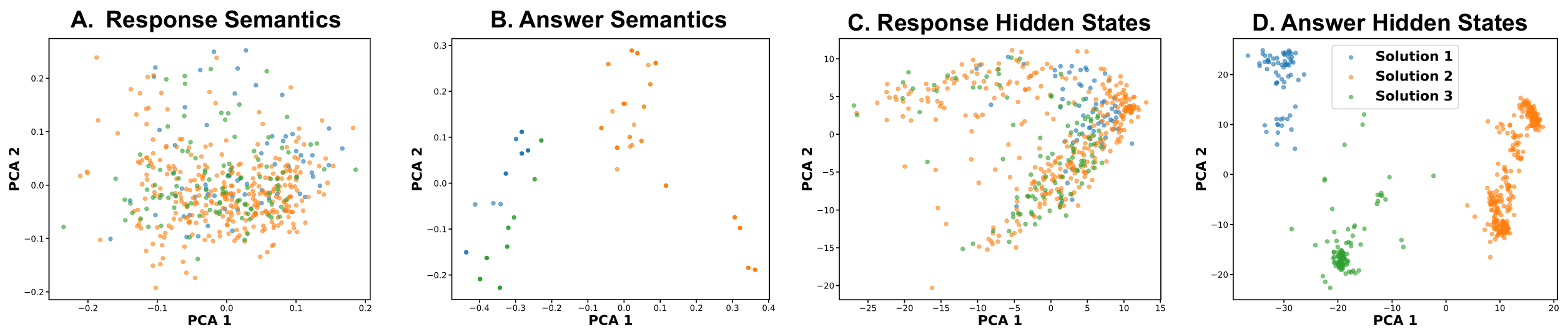}
    \caption{PCA visualization of solution-path clustering for problems (2,6,9,11) across different representation types. The Answer Semantics plots contain fewer points because many responses correspond to identical answers. Results for other problems are provided in Appendix Figure~\ref{fig:clustering_all}.}
    \label{fig:Clustering}
\end{figure*}

\begin{figure}[tbh]
    \centering
    \includegraphics[width=\linewidth]{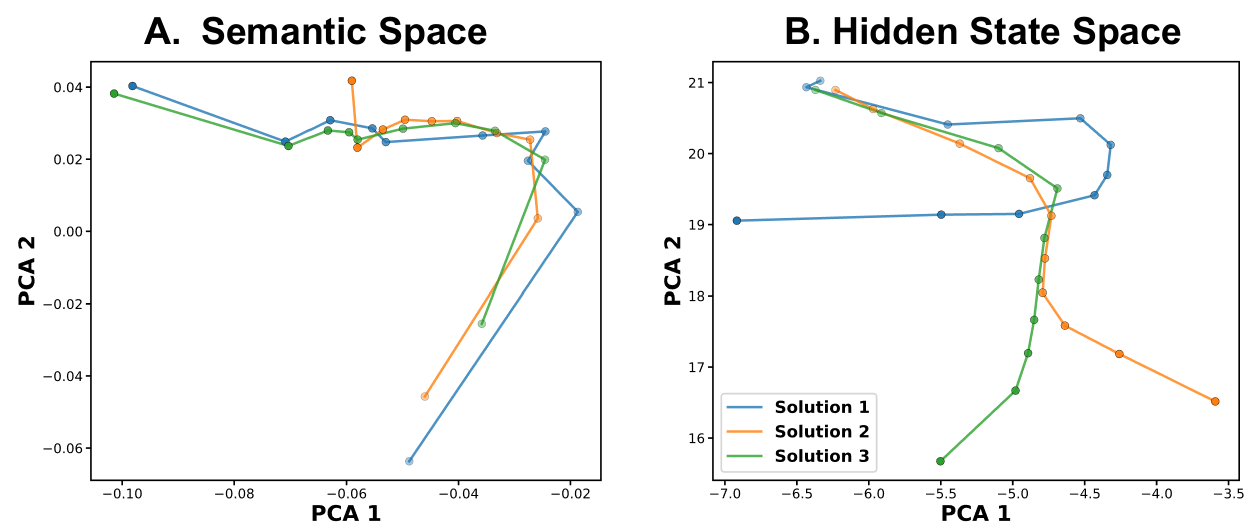}
    \caption{Visualization of solution-path trajectories for problems (2, 6, 9, 11) in semantic and hidden-state spaces. Lighter-colored nodes correspond to earlier prefixes. See Appendix Figure~\ref{fig:trajectory_all} for more results.}
    \label{fig:Trajectory}
\end{figure}

\section{Preliminary Analysis}
\label{sec:preliminary}


In this section, we use Game of 24 as a diagnostic task to study how different representations capture solution diversity. We show that response-level semantic embeddings are often insufficient to distinguish different solution paths, whereas answer-level hidden states provide a more discriminative representation and better reveal how reasoning trajectories diverge during generation.

\subsection{Experimental Setup}

Following \citet{ToT}, we use a multi-solution Game of 24 dataset collected from \url{4nums.com}. It contains 204 problems, each associated with at least four solution categories. Here, a solution category is defined by its underlying mathematical logic: expressions that differ in surface form but follow the same logic (\eg $7*5-10-1$ and $7*5-(10+1)$) are merged according to standard \textit{24-point equivalence rules}\footnote{\url{https://www.4nums.com/theory/}}. 


For each problem $p$, we sample $N=512$ responses $\{r_1, \dots, r_N\}$ from Qwen3-8B~\citep{yang2025qwen3technicalreport}. Each response is a token sequence $r_i = (t_{i,1}, \dots, t_{i,T_i})$, where $T_i$ denotes the response length. We extract the final answer 
and denote its start and end token positions by $s_i$ and $e_i$, respectively. The corresponding answer segment is $a_i = (t_{i,s_i}, \dots, t_{i,e_i})$. Let $h_{i,j}$ denote the hidden state of token $t_{i,j}$.\footnote{We use hidden states from the 32nd layer, which yields the highest separability in our layer-wise analysis (see Appendix~\ref{sec:layer_wise_analysis} for details).} We then compute four representations: response-level semantic embeddings (RespSem), answer-level semantic embeddings (AnsSem), response-level hidden state averages (RespHid), and answer-level hidden state averages (AnsHid):
\[
\begin{array}{l}
\mathrm{RespSem}_i = E_{\mathrm{sem}}(r_i),\\
\mathrm{AnsSem}_i = E_{\mathrm{sem}}(a_i),\\
\mathrm{RespHid}_i = E_{\mathrm{hid}}(r_i) = \frac{1}{T_i}\sum_{j=1}^{T_i} h_{i,j},\\
\mathrm{AnsHid}_i = E_{\mathrm{hid}}(a_i) = \frac{1}{e_i-s_i+1}\sum_{j=s_i}^{e_i} h_{i,j}.
\end{array}
\]
where $E_{\mathrm{sem}}(\cdot)$ denotes Qwen3-Embedding-0.6B~\citep{qwen3embedding}, and $E_{\mathrm{hid}}(\cdot)$ denotes mean pooling over token hidden states.

\subsection{Answer-Level Representations Better Separate Solution Categories}
We first examine whether answer-level representations better separate responses from different solution categories. 
Specifically, for each problem and representation type, we fit a 2D PCA transform on the corresponding representations and visualize their clustering patterns with respect to solution categories.
As shown in Figure~\ref{fig:Clustering}A and \ref{fig:Clustering}C, response-level representations are heavily overlapped and fail to distinguish different solution paths. In contrast, answer-level representations (Figure~\ref{fig:Clustering}B and \ref{fig:Clustering}D) form clear clusters corresponding to each solution category. 
This visual observation is supported by quantitative analysis in Appendix~\ref{sec:quantitative_analysis}: using pairwise AUROC to measure solution separability, answer-level representations outperform response-level representations by 0.36 on average (0.94 vs.\ 0.58).
These results indicate that semantic embeddings of full responses are less effective at distinguishing different solution paths. One plausible explanation is the high structural and stylistic homogeneity among responses to the same problem, which results in high semantic similarity (average cosine similarity of 0.95) and consequently limits response-level representation separability.

\begin{figure*}[t]
    \centering
    \includegraphics[width=0.9\linewidth]{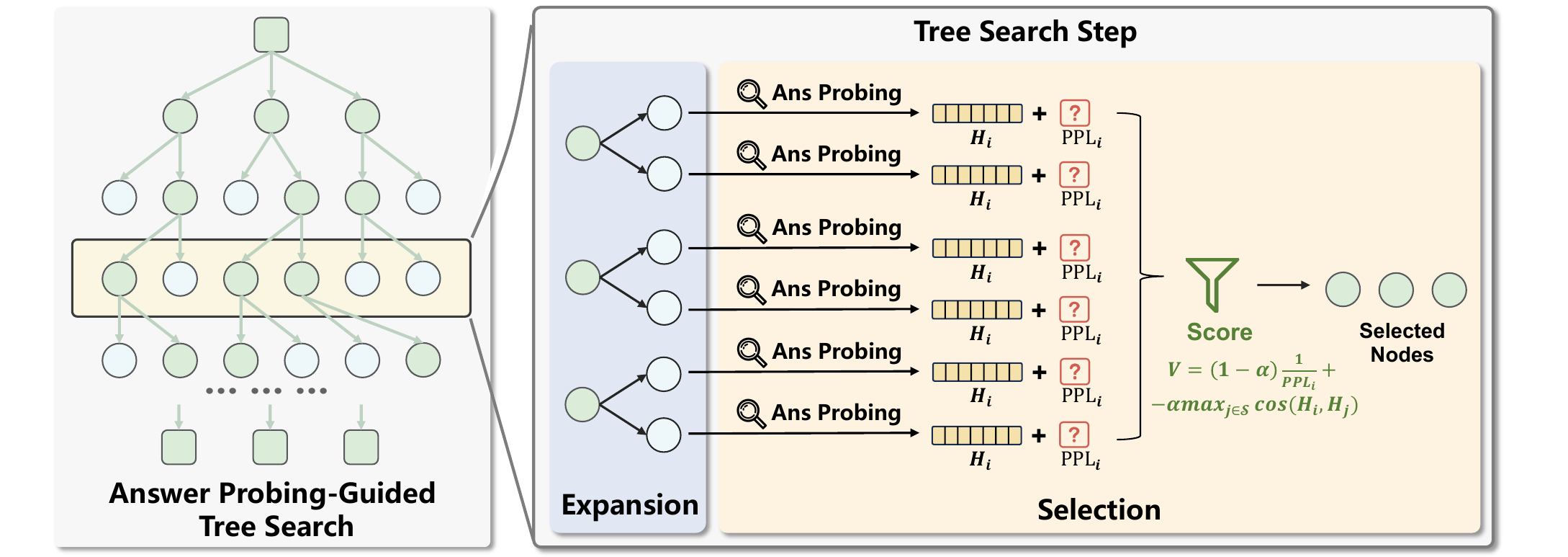}
    \caption{Overview of APTS.}
    \label{fig:apts_pipeline}
\end{figure*}

\subsection{Hidden State Space Better Reveals Trajectory Divergence}
We further analyze how different representation spaces capture trajectory divergence over the course of generation.
Specifically, for each response $r_i = (t_{i,1}, \dots, t_{i,T_i})$, we identify the token positions corresponding to 10\%, 20\%, \dots, 100\% of its length, and use them to construct 10 cumulative prefixes:
\[
r_i^{(k)} = (t_{i,1}, \dots, t_{i,\lfloor kT_i/10 \rfloor}), \qquad k=1,\dots,10.
\]
We then group these prefixes by solution category and average their representations, yielding the semantic and hidden state representations of each solution at 10\%, 20\%, \dots, 100\% of the reasoning process:
\[
\begin{array}{l}
s_{c,\mathrm{sem}}^{(k)} = \frac{1}{|R_c|}\sum_{r_i \in R_c} E_{\mathrm{sem}}(r_i^{(k)}),\\
s_{c,\mathrm{hid}}^{(k)} = \frac{1}{|R_c|}\sum_{r_i \in R_c} E_{\mathrm{hid}}(r_i^{(k)})
\end{array}
\]
where $R_c$ denotes the set of responses belonging to solution category $c$. For trajectory visualization, we fit PCA on answer-level representations (AnsSem or AnsHid) and project the reasoning trajectories (($s_{c,\mathrm{sem}}^{(k)}$ or $s_{c,\mathrm{hid}}^{(k)}$)) into this space to study how reasoning paths progress toward the final answer.

Figure~\ref{fig:Trajectory} shows that the hidden state space yields clearer separation among solution trajectories. Trajectories associated with different solutions originate from nearby regions and progressively diverge as generation proceeds, eventually reaching distinct endpoints. In contrast, trajectories in the semantic space are less coherent: they do not exhibit a clear common origin and remain highly overlapped throughout most of the generation process. Overall, these results suggest that hidden state representations provide a more discriminative view of intermediate reasoning development than semantic embeddings. This is consistent with prior work~\citep{fang2026controllablellmreasoningsparse,reasong_strength}, which shows that hidden state representations capture information about reasoning progress beyond semantic content alone. This finding justifies our use of hidden-state representations instead of semantic embeddings for Answer Probing.


%% file: chapters/3_method.tex
\section{Method}
We present our method in two parts. First, motivated by the finding that answer-level hidden states better distinguish distinct solution paths, we propose \textit{Answer Probing}, which characterizes an intermediate reasoning path by probing the answer that the LLM would generate from that path (Section~\ref{sec:ans_probing}). Second, we introduce APTS, which uses probed answers’ hidden state similarity and PPL to guide tree search, thereby improving solution diversity in LLM generation (Section~\ref{sec:apts}).

\subsection{Answer Probing} 
\label{sec:ans_probing}
As shown in Section~\ref{sec:preliminary}, answer-level hidden states provide more discriminative representations of different solution paths than response-level semantic embeddings. However, during generation, the final answer is unavailable for an intermediate reasoning path. To address this issue, we introduce \textit{Answer Probing}, which characterizes an intermediate reasoning path by probing the answer that the LLM would generate from that path.

Formally, given a problem $p$, let $r$ denote an intermediate reasoning path, represented as a partially generated response. Answer Probing constructs a probing context $c = p \oplus r \oplus \pi_{\text{ans}}$ by appending an answer-inducing prompt $\pi_{\text{ans}}$ (\eg ``The final answer is'')\footnote{We analyze the sensitivity to the probing prompt formulation in Appendix~\ref{sec:probing_prompt_ablation}.} to $r$, and then prompting the LLM to generate a probed answer $PA$:\footnote{We use greedy decoding here to eliminate sampling randomness.}
\[
PA = \operatorname{LLM}(c) = (a_1, a_2, \dots, a_L),
\]
where $L$ is the length of the probed answer. We then use the hidden states of the probed-answer tokens to represent the reasoning path. Specifically, let $h_i$ denote the hidden state of token $a_i$ in $PA$. The representation of $r$ is computed as:
\[
H(r) = \frac{1}{L}\sum_{i=1}^{L} h_i.
\]
By forcing the LLM to produce an answer directly from an intermediate reasoning path, Answer Probing exposes the latent solution tendency encoded in that state. As a result, two intermediate reasoning states that are linguistically similar but lead to different solutions can still be distinguished by their probed-answer representations.

An additional benefit of Answer Probing is that it provides a quality signal. For a probed answer $PA = (a_1, a_2, \dots, a_L)$, we compute its PPL under the current reasoning path as
\[
\mathrm{PPL}(PA)
= \exp\!\left(
-\frac{1}{L}\sum_{i=1}^{L}\log p_\theta(a_i \mid c, a_{<i})
\right).
\]
A lower probed-answer PPL indicates that the LLM produces the answer more confidently from the current path, and thus serves as a useful proxy for the quality of that reasoning path.\footnote{We analyze the distribution of inverse-PPL values across tasks in Appendix~\ref{sec:inverse_ppl_analysis}.}

Overall, Answer Probing provides two useful signals: hidden-state similarity between probed answers, which reflects similarity between underlying solution paths, and probed-answer PPL, which reflects the quality of the current reasoning state.

\subsection{Answer Probing-Guided Tree Search} 
\label{sec:apts}


Building on the two signals provided by Answer Probing, we propose APTS, which guides tree search toward high-quality and diverse candidate nodes, thereby improving solution diversity in LLM generation.

As shown in Figure~\ref{fig:apts_pipeline}, APTS adopts a breadth-first tree search in which each node represents an intermediate reasoning step.\footnote{In this work, reasoning steps are separated using ``\textbackslash n\textbackslash n'' as the delimiter.} 
For a problem $p$ requiring $M$ responses, APTS first samples $M$ initial nodes, denoted by $\mathcal{N}^{(0)} = \{n_1^{(0)}, n_2^{(0)}, \dots, n_M^{(0)}\}$. 
At each search step $d$, APTS performs \textit{Expansion} and \textit{Selection}. 
During the expansion stage, each node in $\mathcal{N}^{(d)}$ generates $W$ child nodes, where $W$ is the node expansion width. This yields a candidate set $\mathcal{C}^{(d+1)} = \{c_1^{(d+1)}, c_2^{(d+1)}, \dots, c_{MW}^{(d+1)}\}$.
During the selection stage, APTS chooses $M$ nodes from $\mathcal{C}^{(d+1)}$ to form $\mathcal{N}^{(d+1)}$ for further exploration. 
The search terminates when the maximum search depth $D$ is reached or all nodes in $\mathcal{N}^{(d)}$ have produced final answers.

To select $\mathcal{N}^{(d+1)}$, APTS performs $M$ rounds of greedy diversity-aware selection while maintaining a selected set $\mathcal{S}$, initially empty. In each round, for each remaining candidate $c_i^{(d+1)} \in \mathcal{C}^{(d+1)} \setminus \mathcal{S}$, we compute the node value as:
\[
\begin{array}{l}
V(c_i^{(d+1)}) = (1-\alpha) \cdot \frac{1}{\mathrm{PPL}_i} + \alpha \cdot DS_i, \\[4pt]
DS_i =
\begin{cases}
0, & \text{if } \mathcal{S} = \emptyset,\\
-\max\limits_{c_j \in \mathcal{S}} \cos(H_i, H_j), & \text{otherwise},
\end{cases}
\end{array}
\]
where the first term measures node quality, the second term $DS_i$ encourages diversity with respect to the set of already selected nodes, and $\alpha \in [0,1]$ is a trade-off coefficient. In each round, the candidate with the highest node value is added to $\mathcal{S}$. After $M$ rounds, $\mathcal{S}$ becomes $\mathcal{N}^{(d+1)}$. Full pseudocode of APTS is provided in Appendix, Algorithm~\ref{alg:apts}.

%% file: chapters/4_experiments.tex
\section{Experiments}

\begin{table*}[t]
\centering
\resizebox{\textwidth}{!}{%
\footnotesize
\setlength{\tabcolsep}{5pt}
\renewcommand{\arraystretch}{1.08}

\begin{tabular}{llcc|cc|ccc}
\toprule
\textbf{Model} & \textbf{Method} 
& \multicolumn{2}{c}{\textbf{Game of 24}}
& \multicolumn{2}{c}{\textbf{Forward Synthesis}}
& \multicolumn{3}{c}{\textbf{TestCaseGen}} \\
\cmidrule(lr){3-4} \cmidrule(lr){5-6} \cmidrule(lr){7-9}
& & \textbf{Acc@16} & \textbf{Cov@16} & \textbf{Conf@16} & \textbf{NCircle@16} & \textbf{Acc@16} & \textbf{L-Cov@16} & \textbf{B-Cov@16} \\
\midrule

\multirow{8}{*}{Qwen3-8B}
& \hspace{0.5em}Repeated Sampling (T=0.5) & 0.9479 & \cellcolor{gray!12}0.6080 & \underline{0.2820} & \cellcolor{gray!12}7.275 & 0.7360 & \cellcolor{gray!12}0.6889 & \cellcolor{gray!12}0.5744 \\
& \hspace{0.5em}Repeated Sampling (T=1.0) & \textbf{0.9559} & \cellcolor{gray!12}0.6179 & \textbf{0.2823} & \cellcolor{gray!12}7.630 & \underline{0.7394} & \cellcolor{gray!12}0.7056 & \cellcolor{gray!12}0.5896 \\
& \hspace{0.5em}Repeated Sampling (T=1.5) & 0.9289 & \cellcolor{gray!12}0.6250 & 0.1513 & \cellcolor{gray!12}7.130 & 0.6138 & \cellcolor{gray!12}0.6859 & \cellcolor{gray!12}0.5731 \\
& \hspace{0.5em}Diverse Beam Search         & 0.9311 & \cellcolor{gray!12}0.6238 & 0.2784 & \cellcolor{gray!12}7.775 & 0.7172 & \cellcolor{gray!12}0.7063 & \cellcolor{gray!12}0.5957 \\
& \hspace{0.5em}SemDiD (NeurIPS'25)                    & \underline{0.9553} & \cellcolor{gray!12}0.6337 & 0.2782 & \cellcolor{gray!12}\underline{7.930} & \textbf{0.7484} & \cellcolor{gray!12}0.7188 & \cellcolor{gray!12}0.6088 \\
& \hspace{0.5em}GuidedSampling (ICLR'26)              & 0.9240 & \cellcolor{gray!12}\underline{0.6383} & 0.2783 & \cellcolor{gray!12}7.715 & 0.6956 & \cellcolor{gray!12}\underline{0.7192} & \cellcolor{gray!12}\underline{0.6106} \\
& \hspace{0.5em}STARS (ICLR'26)                    & 0.9525 & \cellcolor{gray!12}0.6185 & 0.2794 & \cellcolor{gray!12}7.800 & 0.7178 & \cellcolor{gray!12}0.7118 & \cellcolor{gray!12}0.6079 \\
& &  & \cellcolor{gray!12} &  & \cellcolor{gray!12} &  & \cellcolor{gray!12} & \cellcolor{gray!12} \\[-1.0em]
& \hspace{0.5em}\textbf{APTS(Ours)}    & 0.9445 & \cellcolor{gray!12}\textbf{0.6913} & 0.2808 & \cellcolor{gray!12}\textbf{8.150} & 0.7322 & \cellcolor{gray!12}\textbf{0.7306} & \cellcolor{gray!12}\textbf{0.6232} \\
\midrule

\multirow{8}{*}{GPT-OSS-120B}
& \hspace{0.5em}Repeated Sampling (T=0.5) & \textbf{0.9507} & \cellcolor{gray!12}0.6046 & 0.4411 & \cellcolor{gray!12}3.190 & \textbf{0.9347} & \cellcolor{gray!12}0.7965 & \cellcolor{gray!12}0.6927 \\
& \hspace{0.5em}Repeated Sampling (T=1.0) & \underline{0.9292} & \cellcolor{gray!12}0.6211 & \underline{0.4423} & \cellcolor{gray!12}3.255 & \underline{0.9319} & \cellcolor{gray!12}0.8150 & \cellcolor{gray!12}0.7197 \\
& \hspace{0.5em}Repeated Sampling (T=1.5) & 0.9032 & \cellcolor{gray!12}\underline{0.6695} & 0.2483 & \cellcolor{gray!12}2.420 & 0.8753 & \cellcolor{gray!12}\underline{0.8280} & \cellcolor{gray!12}0.7363 \\
& \hspace{0.5em}Diverse Beam Search         & 0.8986 & \cellcolor{gray!12}0.6319 & 0.4369 & \cellcolor{gray!12}3.290 & 0.9297 & \cellcolor{gray!12}0.8242 & \cellcolor{gray!12}0.7335 \\
& \hspace{0.5em}SemDiD (NeurIPS'25)                   & 0.9139 & \cellcolor{gray!12}0.6544 & 0.4333 & \cellcolor{gray!12}\underline{3.330} & 0.9294 & \cellcolor{gray!12}0.8259 & \cellcolor{gray!12}\underline{0.7368} \\
& \hspace{0.5em}GuidedSampling (ICLR'26)    & 0.8879 & \cellcolor{gray!12}0.6432 & 0.4214 & \cellcolor{gray!12}3.300 & 0.8941 & \cellcolor{gray!12}0.8234 & \cellcolor{gray!12}0.7322 \\
& \hspace{0.5em}STARS (ICLR'26)   & 0.9124 & \cellcolor{gray!12}0.6344 & 0.4335 & \cellcolor{gray!12}3.270 & 0.9281 & \cellcolor{gray!12}0.8216 & \cellcolor{gray!12}0.7324 \\
& &  & \cellcolor{gray!12} &  & \cellcolor{gray!12} &  & \cellcolor{gray!12} & \cellcolor{gray!12} \\[-1.0em]
& \hspace{0.5em}\textbf{APTS(Ours)}    & 0.9053 & \cellcolor{gray!12}\textbf{0.6926} & \textbf{0.4499} & \cellcolor{gray!12}\textbf{3.665} & 0.9309 & \cellcolor{gray!12}\textbf{0.8335} & \cellcolor{gray!12}\textbf{0.7430} \\
\bottomrule

\end{tabular}
}
\caption{Comparison of different decoding methods on Game of 24, Forward Synthesis, and TestCaseGen. Bold indicates the best result and underline indicates the second best, and gray background denotes diversity metrics.}
\label{tab:main_results}
\end{table*}

In this section, we conduct experiments to address the following research questions:
\begin{itemize}[itemsep=1pt, topsep=2pt, parsep=0pt, leftmargin=*]
    \item \textbf{RQ1:} How effectively does APTS enhance solution diversity compared to baseline methods?
    \item \textbf{RQ2:} What roles do the two signals in Answer Probing play in APTS, and is AnsHid a more effective representation than alternative representations for APTS?
    \item \textbf{RQ3:} How sensitive is APTS to variations in its key hyperparameters?
\end{itemize}

\subsection{Experiment Setup}

\paragraph{Datasets and Evaluation.}
We conduct experiments on three datasets spanning math, chemistry, and code: the multi-solution Game of 24 dataset introduced in Section~\ref{sec:preliminary}, Forward Synthesis~\citep{yu2024llasmol}, and TestCaseGen~\citep{huang2025benchmarking}. We evaluate on Qwen3-8B~\citep{yang2025qwen3technicalreport} and GPT-OSS-120B~\citep{openai2025gptoss120bgptoss20bmodel}, sampling 16 responses per problem. For Game of 24, we report accuracy (\textbf{Acc@16}) and solution category coverage (\textbf{Cov@16}). For Forward Synthesis, we report product quality (\textbf{Conf@16})~\citep{schwaller2021extraction} and chemical space coverage (\textbf{NCircle@16})~\citep{xie2021much}. For TestCaseGen, we report test accuracy (\textbf{Acc@16}), line coverage (\textbf{L-Cov@16}), and branch coverage (\textbf{B-Cov@16}). 

We use a task-specific diversity metric for each task rather than a single unified metric, because ``genuinely distinct solutions'' means different things across domains: distinct arithmetic structures in math, diverse chemical scaffolds in chemistry, and different program paths in code. Concretely, \textbf{Cov@16} $\in[0,1]$ is the fraction of valid solution categories covered by the 16 answers; \textbf{NCircle@16} $\in[1,16]$ is the size of the largest subset of generated molecules that are mutually dissimilar (Tanimoto similarity below 0.75)~\citep{xie2021much}; and \textbf{L-Cov@16} and \textbf{B-Cov@16} $\in[0,1]$ are the line and branch coverage of the target code achieved by the correct generated tests. For all diversity metrics, higher is better. Each metric is adopted from its task's original benchmark and is community-standard, consistent with prior work on diverse generation~\citep{STARS}. See Appendix~\ref{sec:eval_details} for more details.

\paragraph{Baselines and Implementation.}
We compare APTS with five representative methods spanning decoding-based, search-based, and steering-based approaches, including recent methods from NeurIPS 2025 and ICLR 2026: Repeated Sampling~\citep{renze2024effect}, Diverse Beam Search~\citep{DiverseBeam}, SemDiD~\citep{SemDiD}, GuidedSampling~\citep{GuidedSampling}, and STARS~\citep{STARS}. Baseline and experimental details are provided in Appendix~\ref{sec:impl_details}.

\subsection{Diversity Performance Compared with Baselines (RQ1)}

\paragraph{Main Results.}
We evaluate the performance of different methods in Table~\ref{tab:main_results}, from which we draw the following observations: (1) \textbf{APTS achieves the best solution diversity across all three tasks and both LLMs.} Compared with Repeated Sampling at the same temperature ($T=1.0$), APTS yields an average absolute improvement of over 7\% in Cov@16 on Game of 24 across the two LLMs. On TestCaseGen, it yields an average absolute improvement of over 5\% in the mean of L-Cov@16 and B-Cov@16. On Forward Synthesis, it improves NCircle@16 by 0.47 on average.
These results demonstrate the effectiveness and robustness of APTS; (2) \textbf{Simply increasing the sampling temperature does not necessarily improve the diversity of valid solutions.} In fact, raising the temperature to $T=1.5$ yields lower diversity than $T=1.0$ on Forward Synthesis for both LLMs and on TestCaseGen for Qwen3-8B. This indicates that relying solely on sampling randomness is insufficient for increasing valid solution diversity, because excessive randomness yields superficially diverse yet incorrect outputs; 
(3) \textbf{APTS achieves a better trade-off between correctness and diversity.} Increasing the temperature to $T=1.5$ improves diversity but typically causes a substantial drop in correctness (9\% on average). In contrast, APTS improves diversity while incurring only a negligible correctness decrease of 0.6\% relative to Repeated Sampling ($T=1.0$). This is because APTS explicitly evaluates the quality of candidate paths during search and selects promising paths that maintain both correctness and diversity.

\begin{figure*}[tbh]
    \centering
    \includegraphics[width=0.9\linewidth]{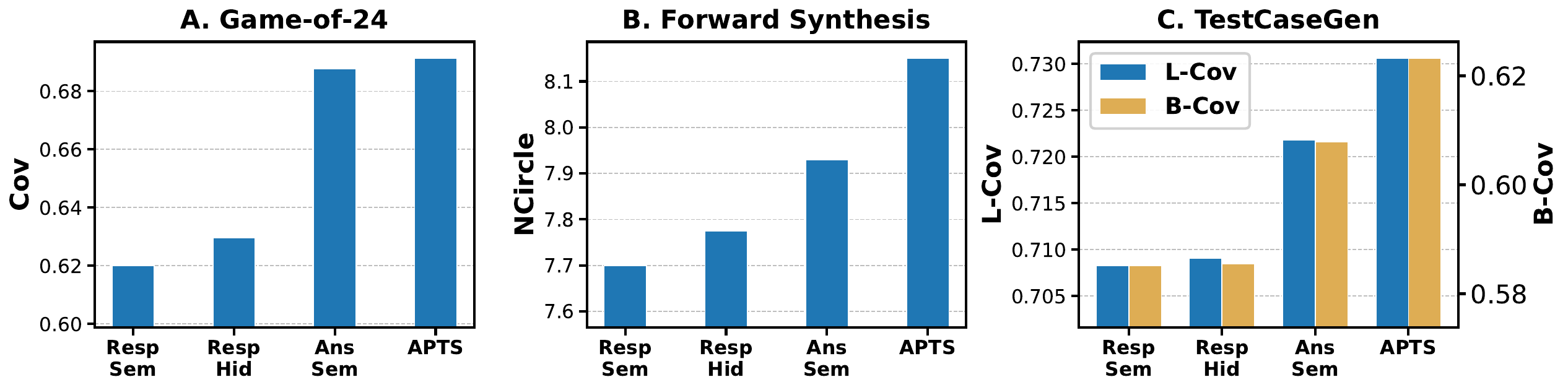}
    \caption{Comparison of different representations used as diversity signals in APTS on Qwen3-8B.}
    \label{fig:Ablation}
\end{figure*}

\paragraph{APTS on Open-Ended Tasks and Closed-Source LLMs.}
To further evaluate the generalizability of APTS, we conduct two additional experiments. First, we evaluate APTS on LiveIdeaBench~\citep{ruan2026evaluating}, an open-ended scientific idea generation benchmark, and find that APTS consistently improves diversity while maintaining quality (see Appendix~\ref{sec:liveideabench}). Second, we adapt APTS to the closed-source GPT-o3-mini by replacing hidden-state signals with semantic embeddings and semantic entropy, and observe consistent diversity improvements (see Appendix~\ref{sec:closed_source}). These results show that APTS can be effectively applied to open-ended generation tasks and adapted to closed-source LLMs.

\paragraph{Latency and Efficiency Analysis.}
Considering that APTS introduces additional computational overhead due to tree search and answer probing, we measure the wall-clock latency of APTS. Taking Game of 24 on Qwen3-8B as an example, APTS introduces only 1.46$\times$ overhead over Repeated Sampling with the same sample count ($N\!=\!16$), thanks to prefix caching and batching of short continuations. We also measure the latency of budget-matched Repeated Sampling, where the number of samples is increased until its total token cost matches that of APTS ($N\!=\!29$ for Game of 24), and find that APTS ($N\!=\!16$) is faster than budget-matched Repeated Sampling ($N\!=\!29$). Moreover, we compare the solution diversity under this budget-matched setting and find that APTS still achieves higher diversity than Repeated Sampling with more trials. This further demonstrates the effectiveness of APTS. See Appendix~\ref{sec:wall_clock_analysis} for details.

\subsection{Understanding the Role of Answer Probing in APTS (RQ2)}

\paragraph{Effectiveness of probing-answer hidden states as diversity signals.} 
We further assess the effectiveness of using probing-answer hidden states as the diversity signal for APTS by comparing it with alternative representations.
Specifically, we replace it with alternative representations introduced in Section~2, including RespSem, AnsSem, and RespHid, and use them to compute the similarity penalty that guides node selection in APTS. The results are shown in Figure~\ref{fig:Ablation}, from which we draw the following observations: 
\textbf{(1)} Answer-level representations consistently achieve higher solution diversity than response-level representations across all three tasks. This result is consistent with our analysis in Section~2, suggesting that answer-level representations better capture distinctions among solutions; and \textbf{(2)} Compared with AnsSem, APTS using the hidden states of probed answers as the diversity signal achieves clear improvements on TestCaseGen and Forward Synthesis. This indicates that the hidden-state space is more discriminative than the semantic embedding space, allowing APTS to better distinguish different reasoning trajectories.

\begin{figure}[t]
    \centering
    \includegraphics[width=0.85\linewidth]{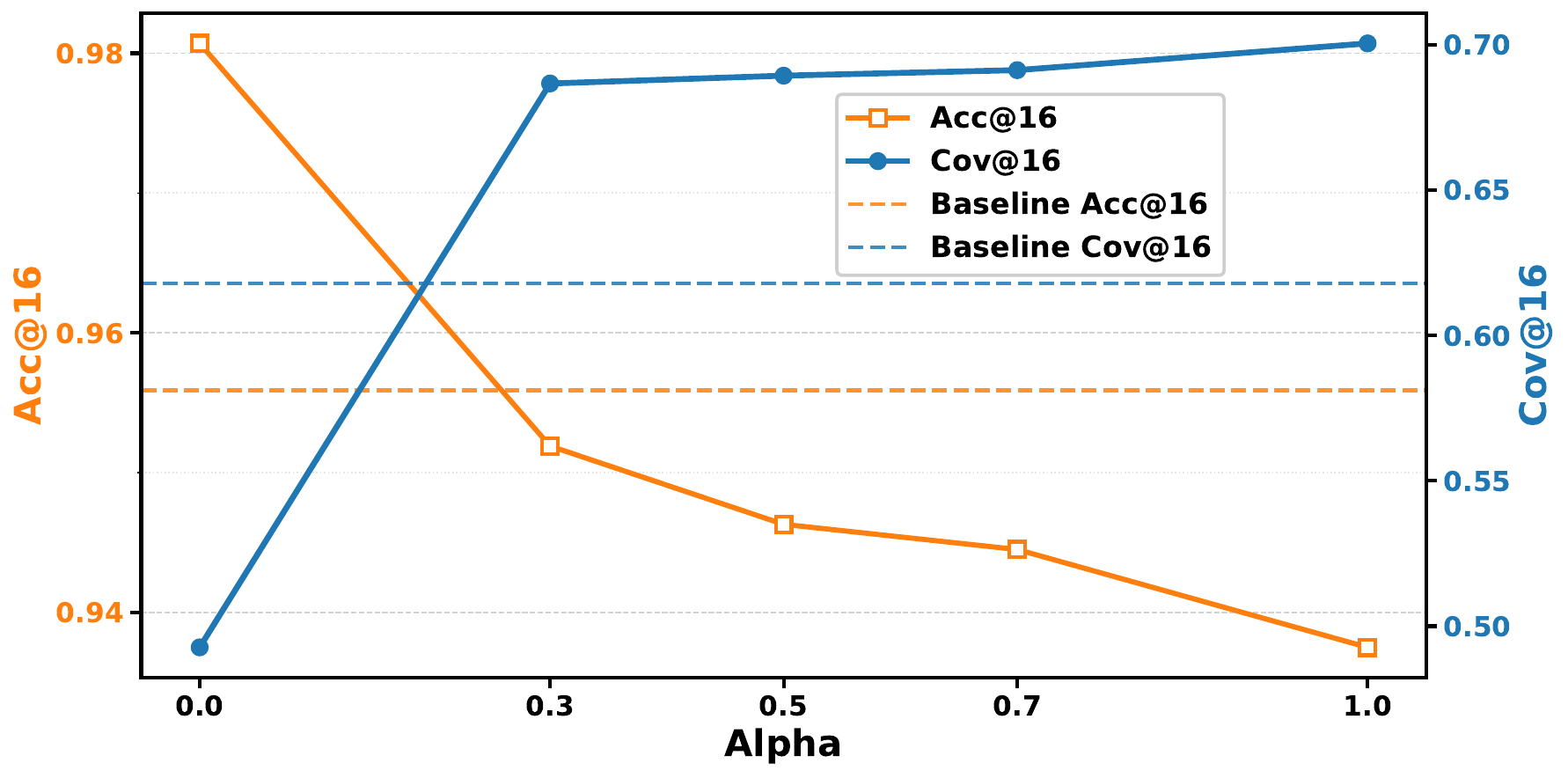}
    \caption{Effect of the trade-off coefficient $\alpha$ on Game of 24 with Qwen3-8B. Results for other datasets are provided in Appendix Figure~\ref{fig:alpha_appendix}.}
    \label{fig:alpha_analysis}
\end{figure}

\paragraph{Roles of the two Answer Probing signals in APTS.} 
To understand the respective roles of the two Answer Probing signals, we evaluate APTS under different values of the trade-off coefficient $\alpha$. 
Figure~\ref{fig:alpha_analysis} presents the results on Game of 24 with Qwen3-8B, and we observe the same trend on TestCaseGen and Forward Synthesis (Appendix Figure~\ref{fig:alpha_appendix}). From these results, we draw the following observations: (1) When $\alpha=0$, APTS selects nodes solely based on probed-answer PPL. In this setting, APTS achieves higher accuracy than Repeated Sampling (0.9807 vs.\ 0.9559), demonstrating that probed-answer PPL is an effective signal for estimating reasoning correctness; and (2) As $\alpha$ increases, the hidden-state similarity penalty in Answer Probing plays a larger role in node selection, leading to steadily improved solution coverage. Notably, once the similarity penalty is introduced (\ie $\alpha \neq 0$), APTS consistently achieves higher solution coverage than the baseline, demonstrating the effectiveness of hidden-state similarity for encouraging diverse exploration.


\subsection{Impact of Hyperparameters (RQ3)}

\begin{figure}[t]
    \centering
    \includegraphics[width=\linewidth]{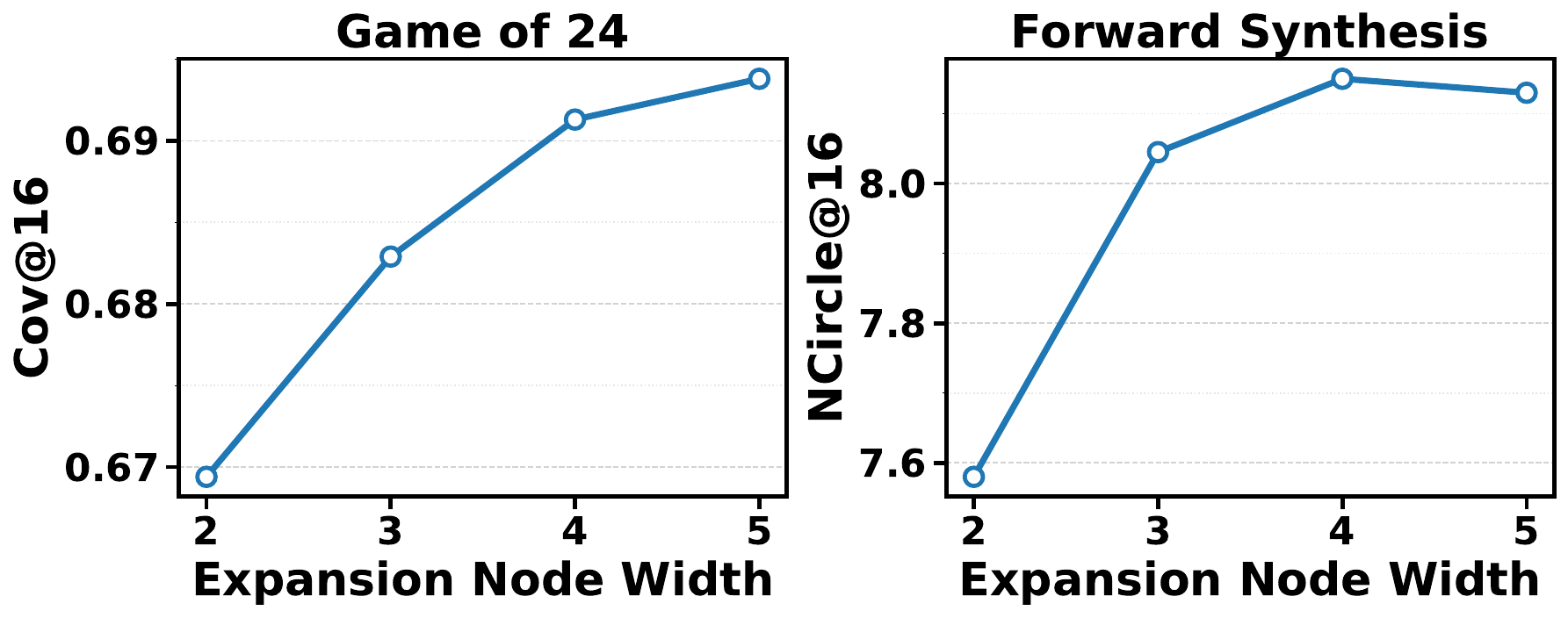}
    \caption{Effect of node expansion width on the solution diversity of APTS on Qwen3-8B.}
    \label{fig:nbw_analysis}
\end{figure}

\paragraph{Influence of Node Expansion Width.}
We study how the node expansion width of APTS affects solution diversity. As shown in Figure~\ref{fig:nbw_analysis}, increasing the expansion width from 2 to 4 leads to substantial improvements in solution diversity. However, when the width is further increased from 4 to 5, the gains become much smaller on Game of 24 (with the same trend also observed on TestCaseGen; see Appendix Figure~\ref{fig:nbw_code}), and even slightly decrease on Forward Synthesis. These results suggest that setting the node expansion width to 4 provides a good trade-off. Larger widths incur additional reasoning overhead while bringing only marginal diversity improvements.

\paragraph{Influence of Max Search Depth.}
We also study how the maximum search depth of APTS affects solution diversity. As shown in Figure~\ref{fig:depth_analysis}, we observe two distinct trends across tasks. For Game of 24, solution diversity increases steadily as the search depth grows. In contrast, for Forward Synthesis, solution diversity improves notably as the depth increases up to 15, but further increasing the depth beyond 15 brings only marginal gains (we observe a similar trend on TestCaseGen; see Appendix Figure~\ref{fig:nbw_depth}). This suggests that the effect of search depth is task-dependent.

We think this may be because, for Game of 24, different solution paths may diverge relatively late, so deeper search can continue to uncover new valid solutions. However, for tasks such as Forward Synthesis and TestCaseGen, most useful branching may occur within a moderate depth, after which the LLM tends to follow previously explored reasoning patterns and produce similar answers.
These results suggest that the optimal search depth should be selected according to the task. In practice, it is preferable to tune the search depth on a validation set and choose the smallest value that captures most of the diversity gains.

%% file: chapters/5_related_work.tex
\section{Related Work}
\paragraph{Diversity Generation of LLMs.} Prior work improves LLM output diversity at inference time mainly through two directions: \emph{prompt-side diversification}, which varies the input prompt~\citep{prompt_diverse_1,GuidedSampling}, and \emph{generation-side diversification}, which modifies the generation process under a fixed prompt. These directions are complementary, and we focus on the latter in this work. Generation-side methods can be further divided into \emph{decoding-based} and \emph{search-based} approaches. Decoding-based methods adjust parameters such as temperature~\citep{renze2024effect} and top-p~\citep{HoltzmanBDFC20}, but often yield only token-level variation. Search-based methods explore multiple reasoning paths and prune similar ones using semantic embeddings~\citep{SemDiD}. However, semantic similarity can be confounded by linguistic or stylistic noise, making truly distinct reasoning paths hard to detect. Answer probing addresses this issue by focusing on answer-level signals from intermediate states.

\begin{figure}[t]
    \centering
    \includegraphics[width=\linewidth]{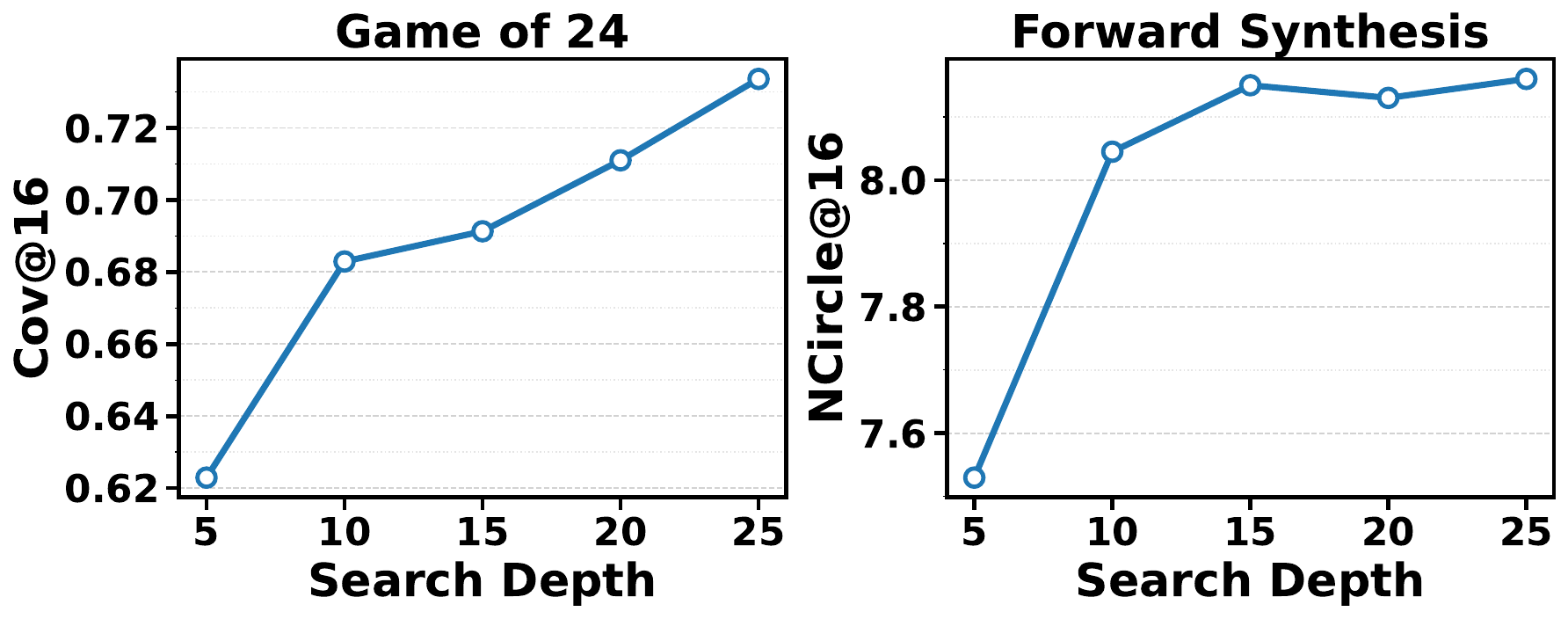}
    \caption{Effect of max search depth on the solution diversity of APTS on Qwen3-8B.}
    \label{fig:depth_analysis}
\end{figure}

\paragraph{Positioning of Answer Probing.}
Several recent works explore mechanisms related to Answer Probing.
Certaindex~\citep{Certaindex} introduces a probe-in-the-middle mechanism that extracts intermediate answers during reasoning to estimate certainty and enable early stopping. Answer Probing extends this probing mechanism to serve as both a diversity signal (via hidden-state similarity between probed answers) and a quality signal (via probed-answer PPL), rather than solely estimating certainty.
On the diversity side, prior work mainly studies diversity in semantic embedding space (\eg SD-E$^2$~\citep{SDE2}), or, when using hidden states, focuses on global trajectory geometry rather than solution-level convergence (\eg VERL~\citep{VERL}). In contrast, Answer Probing defines diversity at the level of underlying solution tendency---whether two intermediate states are likely to converge to the same final solution. As shown in Section~\ref{sec:preliminary}, this representation better separates distinct solution paths than semantic embeddings, which are easily confounded by linguistic and stylistic similarities.
Additionally, tree-structured generation has been adopted for credit assignment in LLM reasoning~\citep{SPOtree} and RL post-training of visual generative models~\citep{TreeGRPO}. While APTS also employs tree search, its contribution lies in the answer-probing-based signals used to guide node selection.

%% file: chapters/6_conclusion.tex
\section{Conclusion}

In this work, we propose Answer Probing, which probes the potential answers an LLM would reach from an intermediate reasoning state. The probed answers provide two useful reasoning signals: the hidden-state similarity between probed answers reflects the diversity of reasoning trajectories, while the PPL of a probed answer reflects the quality of the underlying reasoning. Based on these two signals, we further propose APTS, which guides tree search toward candidate nodes that are both high-quality and diverse, thereby improving solution-level diversity in LLM generation while maintaining strong overall performance.

\section*{Limitations}
While APTS improves solution diversity in LLM generation, it still has several limitations. 
First, APTS requires sampling multiple candidate reasoning paths at each search depth and probing their potential answers, which introduces additional computational overhead. 
Second, APTS mainly focuses on solution diversity rather than semantic diversity, and thus may be less suitable for non-reasoning tasks such as creative writing.
Third, the original APTS formulation depends on access to intermediate hidden states and log probabilities, limiting its direct applicability to closed-source models.
Fourth, the trade-off coefficient $\alpha$ may not transfer reliably across all tasks and models; in practice, a suitable $\alpha$ can be selected using a small validation set.
Fifth, the probed-answer PPL used as a quality signal relies on the model's own confidence, which may be unreliable in rare or challenging scenarios where the model has blind spots.
We provide a detailed case study and error analysis in Appendix~\ref{sec:case_study}.

%% file: chapters/7_appendix.tex
\newpage
\section{Layer-wise Analysis of Hidden-State Separability}
\label{sec:layer_wise_analysis}

To determine which hidden layer provides the best separability across solution categories, we conduct a layer-wise analysis on the Game of 24 dataset using Qwen3-8B. For each problem, we sample 512 responses and extract their answer-level hidden-state representations from every transformer layer. For each solution category, we then compute the mean of all answer-level hidden-state representations within that category and use it as the category-level representation.

\begin{figure}[htb]
    \centering
    \includegraphics[width=\linewidth]{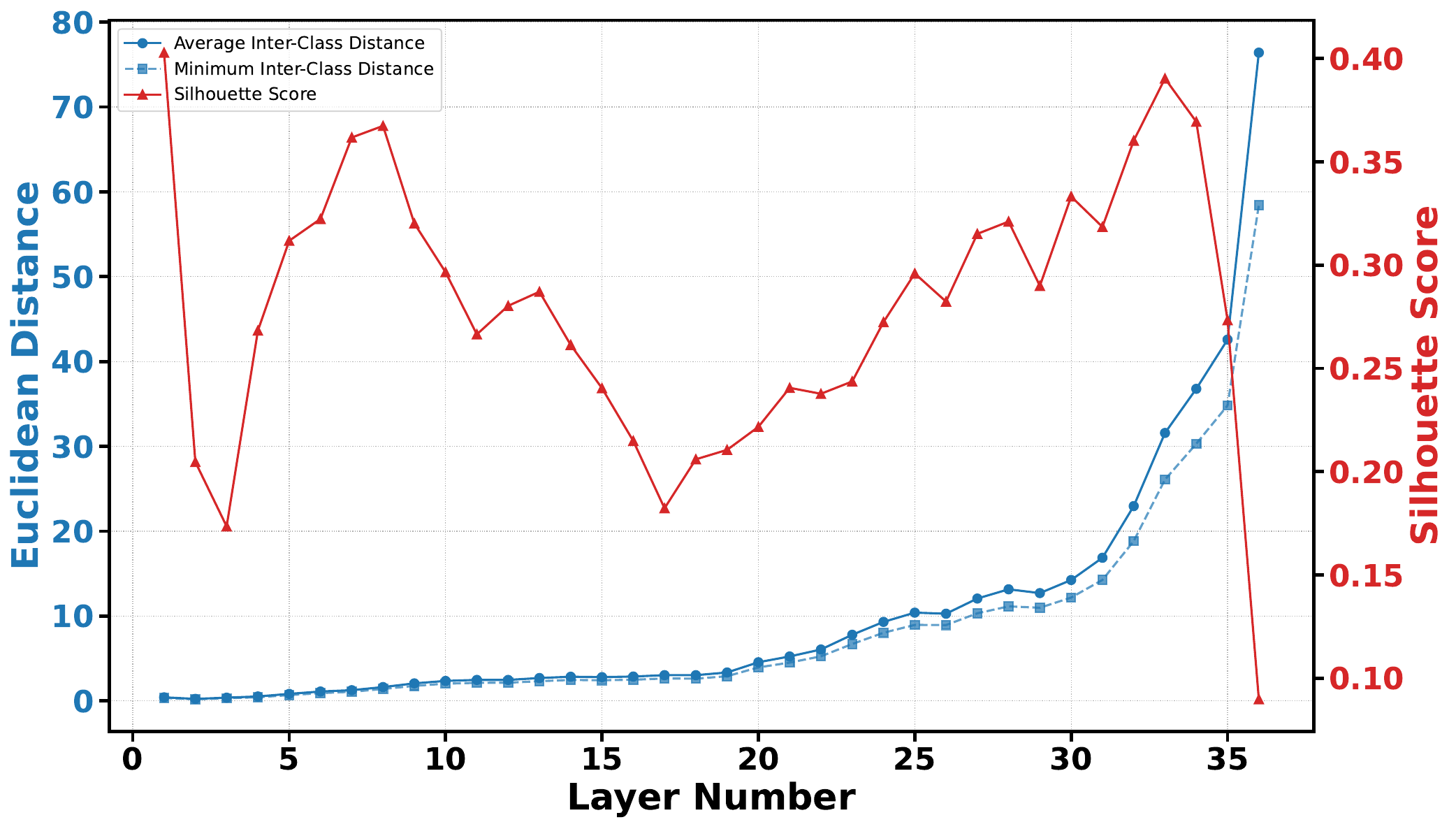}
    \caption{Layer-wise separability analysis of answer-level hidden states on Game of 24 using Qwen3-8B. }
    \label{fig:layer_wise_analysis}
\end{figure}

We quantify how well different solution categories are separated in each layer using three metrics: average inter-class Euclidean distance, minimum inter-class Euclidean distance, and silhouette score. The first two metrics measure the overall and worst-case separation, respectively, among category-level representations. Larger values indicate better separability between different solution categories. 
The silhouette score measures how well individual answer-level hidden-state representations cluster according to their solution categories, jointly reflecting intra-category compactness and inter-category separation. Larger values indicate better clustering quality.
As shown in Figure~\ref{fig:layer_wise_analysis}, the 32nd layer achieves both strong inter-class separation and high clustering quality. We therefore use the 32nd layer in our main experiments.

\section{Quantitative Analysis of Solution Separability}
\label{sec:quantitative_analysis}
To quantitatively evaluate how well different representations separate solution categories, we conduct a pairwise separability analysis. Specifically, for each problem, we randomly sample 10,000 response pairs from the 512 responses generated by Qwen3-8B on the Game of 24 dataset. We compute the cosine similarity for each pair under four representations: RespSem, RespHid, AnsSem and AnsHid. We treat response pairs from the same solution category as positive examples and use their cosine similarity as the prediction score. We then compute the AUROC for distinguishing same-category pairs from different-category pairs under each representation. A higher AUROC indicates that the representation better captures solution-level distinctions. 

Table~\ref{tab:quantitative_separability} shows that answer-level representations substantially outperform response-level representations in terms of solution separability. We believe this is mainly because many responses share the same final answer, resulting in identical answer-level semantic embeddings. This makes same-category response pairs especially easy to distinguish from different-category pairs under AnsSem, thereby slightly improving the AUROC. 

However, for APTS, AnsHid achieves better performance than AnsSem, as shown in Figure~\ref{fig:Ablation}. This suggests that while AnsSem is more effective at separating responses based on their final answers, AnsHid is better at distinguishing differences in their intermediate reasoning processes. We hypothesize that this is because hidden-state representations preserve richer reasoning-related information beyond surface-level answer semantics. Consequently, AnsHid is more advantageous for answer probing.

\begin{table}[htb]
\centering
\small
\caption{Pairwise AUROC for solution separability under different representations on the Game of 24 dataset using Qwen3-8B. Higher is better.}
\label{tab:quantitative_separability}
\begin{tabular}{cccc}
\toprule
\textbf{RespSem} & \textbf{RespHid} & \textbf{AnsSem} & \textbf{AnsHid} \\
\midrule
0.57 & 0.58 & 0.95 & 0.92 \\
\bottomrule
\end{tabular}
\end{table}

\section{Sensitivity to Probing Prompt Formulation}
\label{sec:probing_prompt_ablation}

To examine whether APTS is sensitive to the exact wording of the probing prompt, we conduct experiments on Game of 24 with Qwen3-8B using three different probing prompts: ``Final answer:'', ``Based on the reasoning so far, the answer is'', and ``My current best answer is''.

We first compare the probed-answer hidden states across prompt variants. For 100 randomly selected intermediate reasoning trajectories, the resulting probed-answer hidden states exhibit an average cosine similarity of 0.9516 across prompts, indicating that the hidden-state representation is highly consistent despite changes in prompt wording.

We further evaluate the full APTS pipeline using each probing prompt. As shown in Table~\ref{tab:probing_prompt}, the final Acc@16 and Cov@16 are very close across all three formulations, suggesting that APTS is not highly sensitive to the exact wording of the probing prompt.

\begin{table}[htb]
\centering
\small
\caption{Effect of probing prompt formulation on Game of 24 with Qwen3-8B.}
\label{tab:probing_prompt}
\resizebox{\linewidth}{!}{%
\begin{tabular}{lcc}
\toprule
\textbf{Probing Prompt} & \textbf{Acc@16} & \textbf{Cov@16} \\
\midrule
Final answer & 0.9445 & 0.6913 \\
Based on the reasoning so far, the answer is & 0.9439 & 0.6927 \\
My current best answer is & 0.9463 & 0.6952 \\
\bottomrule
\end{tabular}
}
\end{table}

\section{Inverse-PPL Distribution Analysis}
\label{sec:inverse_ppl_analysis}

One potential concern is that the inverse-PPL term in the node value function is unbounded, while the cosine similarity term is bounded in $[-1,1]$, leading to an unpredictable diversity--quality trade-off. However, because the probed answer is generated via greedy decoding, the token probabilities along the probed answer tend to be relatively high, making the resulting inverse-PPL naturally concentrated and close to 1.

We verify this empirically on Qwen3-8B across all three tasks. As shown in Table~\ref{tab:inverse_ppl}, the actual inverse-PPL values used by APTS are tightly concentrated within roughly $[0.85, 1.00]$ across tasks, and do not vary by orders of magnitude. This suggests that the quality term remains well-behaved in practice, and a fixed $\alpha$ can work reasonably across different tasks.

\begin{table}[htb]
\centering
\small
\caption{Distribution of inverse-PPL values used by APTS on Qwen3-8B.}
\label{tab:inverse_ppl}
\begin{tabular}{lccc}
\toprule
\textbf{Task} & \textbf{Avg} & \textbf{Min} & \textbf{Max} \\
\midrule
Game of 24 (Math) & 0.95 & 0.87 & 1.00 \\
TestCaseGen (Code) & 0.96 & 0.88 & 1.00 \\
Forward Synthesis (Chem) & 0.94 & 0.85 & 1.00 \\
\bottomrule
\end{tabular}
\end{table}

\section{Detailed Evaluation Protocol}
\label{sec:eval_details}

We conduct experiments on three datasets. We select these tasks because they require LLM reasoning to ensure correctness, while solution diversity is also valuable in these settings.

\paragraph{Game of 24.} The multi-solution Game of 24 dataset is introduced in Section~\ref{sec:preliminary}. We report \textbf{Acc@16}, which measures the average correctness of the 16 responses, and \textbf{Cov@16}, which measures coverage over all possible solution categories.

\paragraph{Forward Synthesis.} In forward synthesis prediction~\citep{yu2024llasmol}, LLMs are asked to generate multiple plausible products from given reactants and reagents, so both chemical validity and diversity are crucial. We report \textbf{Conf@16}, which evaluates the quality of generated products using RXNMapper confidence~\citep{schwaller2021extraction}, and \textbf{NCircle@16}~\citep{xie2021much}, which measures the extent of chemical space explored by computing the largest subset of generated molecules such that no two molecules have Tanimoto similarity~\citep{bajusz2015tanimoto} above a predefined threshold 0.75.

\paragraph{TestCaseGen.} In TestCaseGen~\citep{huang2025benchmarking}, the goal is to generate unit tests for code, where more diverse test cases can improve coverage of different program behaviors and edge cases. Following prior work~\citep{wang2025testeval}, we report \textbf{Acc@16}, which measures whether the generated test cases are executable and contain correct test assertions, as well as \textbf{L-Cov@16} and \textbf{B-Cov@16}, which measure the overall line coverage and branch coverage achieved by all correct generated test cases on the target code.

\section{Implementation Details}
\label{sec:impl_details}

\paragraph{Baselines.}
We compare APTS with the following methods:
\begin{itemize}[itemsep=1pt, topsep=2pt, parsep=0pt, leftmargin=*, label=$\circ$]
    \item \textbf{Repeated Sampling}~\citep{renze2024effect}: generates multiple samples using high temperature sampling.
    \item \textbf{Diverse Beam Search}~\citep{DiverseBeam}: uses a tree-search framework and encourages diversity through token-level diversity penalties.
    \item \textbf{SemDiD}~\citep{SemDiD}: adopts a tree-search framework, using semantic embeddings to guide exploration and prune semantically redundant branches.
    \item \textbf{GuidedSampling}~\citep{GuidedSampling}: first generates diverse solution concepts and then samples answers conditioned on them.
    \item \textbf{STARS}~\citep{STARS}: steers hidden activations at inference time to encourage divergent generation.
\end{itemize}

\paragraph{Hyperparameters.}
Following prior work~\citep{SemDiD}, we test temperatures of 0.5, 1.0, and 1.5 for Repeated Sampling. For Diverse Beam Search, SemDiD, and APTS, we set the sampling temperature to 1.0, the node expansion width $W$ to 4, and the maximum search depth $D$ to 15. All other hyperparameters for Diverse Beam Search and SemDiD follow their original papers. For APTS, we set the trade-off coefficient $\alpha = 0.7$.







\section{Evaluation on Open-Ended Generation}
\label{sec:liveideabench}

To evaluate whether APTS generalizes beyond explicit multi-solution settings, we conduct experiments on LiveIdeaBench~\citep{ruan2026evaluating}, a benchmark that requires LLMs to generate scientific ideas from carefully designed keywords. This setting better reflects open-ended generation scenarios where ``solution diversity'' is less well-defined.

Following the original LiveIdeaBench evaluation protocol, both quality and diversity are scored by multiple LLM judges on a 1--10 scale. Quality evaluates the generated ideas in terms of originality, feasibility, and clarity, while diversity measures how different the ideas generated for the same keyword are. For each keyword, we generate 16 ideas.

As shown in Table~\ref{tab:liveideabench}, APTS achieves the best or near-best quality while consistently improving diversity on both models, suggesting that it is also effective for open-ended generation tasks beyond explicit multi-solution settings.

\begin{table}[htb]
\centering
\small
\caption{Results on LiveIdeaBench. Bold indicates the best diversity score.}
\label{tab:liveideabench}
\resizebox{\linewidth}{!}{%
\begin{tabular}{llcc}
\toprule
\textbf{Model} & \textbf{Method} & \textbf{Quality@16} & \textbf{Diverse@16} \\
\midrule
\multirow{4}{*}{Qwen3-8B}
& Repeated Sampling & 6.27 & 4.00 \\
& Diverse Beam Search & 6.11 & 4.08 \\
& SemDiD & 6.23 & 4.13 \\
& \textbf{APTS (Ours)} & 6.25 & \textbf{4.19} \\
\midrule
\multirow{4}{*}{GPT-OSS-120B}
& Repeated Sampling & 6.81 & 6.22 \\
& Diverse Beam Search & 6.76 & 6.31 \\
& SemDiD & 6.79 & 6.37 \\
& \textbf{APTS (Ours)} & 6.82 & \textbf{6.44} \\
\bottomrule
\end{tabular}
}
\end{table}

\section{Adaptation to Closed-Source Models}
\label{sec:closed_source}

The original APTS formulation relies on hidden states and log probabilities, which are typically unavailable for closed-source models. To extend APTS to this setting, we adapt it by (1) replacing hidden-state embeddings with semantic embeddings of probed answers, and (2) replacing PPL with the semantic entropy~\citep{farquhar2024detecting} of probed answers.

We evaluate this adapted variant on GPT-o3-mini. As shown in Table~\ref{tab:closed_source}, the adapted APTS still improves diversity over Repeated Sampling across all three tasks, suggesting that the core idea of answer probing for search guidance transfers beyond open-weight models.

\begin{table}[htb]
\centering
\small
\caption{Adapted APTS on GPT-o3-mini. Bold indicates the best diversity score.}
\label{tab:closed_source}
\resizebox{\linewidth}{!}{%
\begin{tabular}{lccccccc}
\toprule
& \multicolumn{2}{c}{\textbf{Game of 24}} & \multicolumn{2}{c}{\textbf{Forward Synthesis}} & \multicolumn{3}{c}{\textbf{TestCaseGen}} \\
\cmidrule(lr){2-3} \cmidrule(lr){4-5} \cmidrule(lr){6-8}
\textbf{Method} & Acc@16 & Cov@16 & Conf@16 & NCircle@16 & Acc@16 & L-Cov@16 & B-Cov@16 \\
\midrule
Repeated Sampling & 0.9084 & 0.6796 & 0.4092 & 5.23 & 0.9175 & 0.7288 & 0.6120 \\
\textbf{APTS (Ours)} & 0.9105 & \textbf{0.7162} & 0.4033 & \textbf{5.75} & 0.9153 & \textbf{0.7390} & \textbf{0.6329} \\
\bottomrule
\end{tabular}
}
\end{table}

\section{Evaluation on an Additional Model Family}
\label{sec:deepseek}

To assess whether the effectiveness of APTS generalizes across model families, we further evaluate it on DeepSeek-R1-Distill-Llama-8B, a reasoning model from a different family than Qwen3-8B and GPT-OSS-120B. We report results on Game of 24 and Forward Synthesis. As shown in Table~\ref{tab:deepseek}, APTS achieves the best solution diversity across both tasks (Cov@16 and NCircle@16), confirming that answer-probing-guided search transfers across model families. Together with Qwen3-8B, GPT-OSS-120B, and GPT-o3-mini (Appendix~\ref{sec:closed_source}), our evaluation covers four models spanning different families, scales (8B to 120B), and access types (open-weight and closed-source).

\begin{table}[htb]
\centering
\small
\caption{Results on DeepSeek-R1-Distill-Llama-8B. Gray background denotes diversity metrics; bold indicates the best diversity score.}
\label{tab:deepseek}
\resizebox{\linewidth}{!}{%
\begin{tabular}{lcc|cc}
\toprule
& \multicolumn{2}{c}{\textbf{Game of 24}} & \multicolumn{2}{c}{\textbf{Forward Synthesis}} \\
\cmidrule(lr){2-3} \cmidrule(lr){4-5}
\textbf{Method} & \textbf{Acc@16} & \textbf{Cov@16} & \textbf{Conf@16} & \textbf{NCircle@16} \\
\midrule
Repeated Sampling (T=0.5) & 0.5564 & \cellcolor{gray!12}0.5564 & 0.2124 & \cellcolor{gray!12}6.140 \\
Repeated Sampling (T=1.0) & 0.7518 & \cellcolor{gray!12}0.6526 & 0.1710 & \cellcolor{gray!12}5.205 \\
Repeated Sampling (T=1.5) & 0.0064 & \cellcolor{gray!12}0.0199 & 0.0006 & \cellcolor{gray!12}0.035 \\
Diverse Beam Search & 0.7411 & \cellcolor{gray!12}0.6694 & 0.2105 & \cellcolor{gray!12}6.745 \\
SemDiD & 0.7439 & \cellcolor{gray!12}0.6727 & 0.2111 & \cellcolor{gray!12}6.970 \\
\textbf{APTS (Ours)} & 0.7475 & \cellcolor{gray!12}\textbf{0.6927} & 0.2111 & \cellcolor{gray!12}\textbf{7.175} \\
\bottomrule
\end{tabular}%
}
\end{table}

\section{Latency and Efficiency Analysis}
\label{sec:wall_clock_analysis}

\paragraph{Wall-Clock Latency.}
We measure the end-to-end inference latency of APTS on Game of 24 with Qwen3-8B, using vLLM on a single A100-80GB GPU under the same decoding settings as in the main experiments. Both APTS and Repeated Sampling reach a peak GPU memory of 72GB.

\begin{table}[bth]
\centering
\small
\caption{Average per-problem wall-clock time on Game of 24 with Qwen3-8B.}
\label{tab:wall_clock}
\resizebox{\linewidth}{!}{%
\begin{tabular}{lc}
\toprule
\textbf{Method} & \textbf{Wall-clock time} \\
\midrule
Repeated Sampling ($N\!=\!16$) & 8.64s \\
Repeated Sampling (budget-matched, $N\!=\!29$) & 13.90s \\
APTS ($N\!=\!16$) & 12.61s \\
\bottomrule
\end{tabular}%
}
\end{table}

As shown in Table~\ref{tab:wall_clock}, APTS introduces 1.46$\times$ wall-clock overhead over Repeated Sampling with the same sample count, and is faster than budget-matched Repeated Sampling. The latency gap is narrowed by two implementation optimizations:
\begin{itemize}[itemsep=1pt, topsep=2pt, parsep=0pt, leftmargin=*]
    \item \textbf{Prefix caching.} Because tree nodes share long prefixes, both node expansion and answer probing can reuse cached prefixes and only decode the short suffix specific to each node.
    \item \textbf{Batching.} Since both stages generate only short continuations, many tree nodes can be packed into a single LLM generation call. In our setup, we can batch up to 640 expansions in one call on a single A100-80GB GPU without exceeding memory limits.
\end{itemize}

\begin{figure}[tbh]
    \centering
    \includegraphics[width=\linewidth]{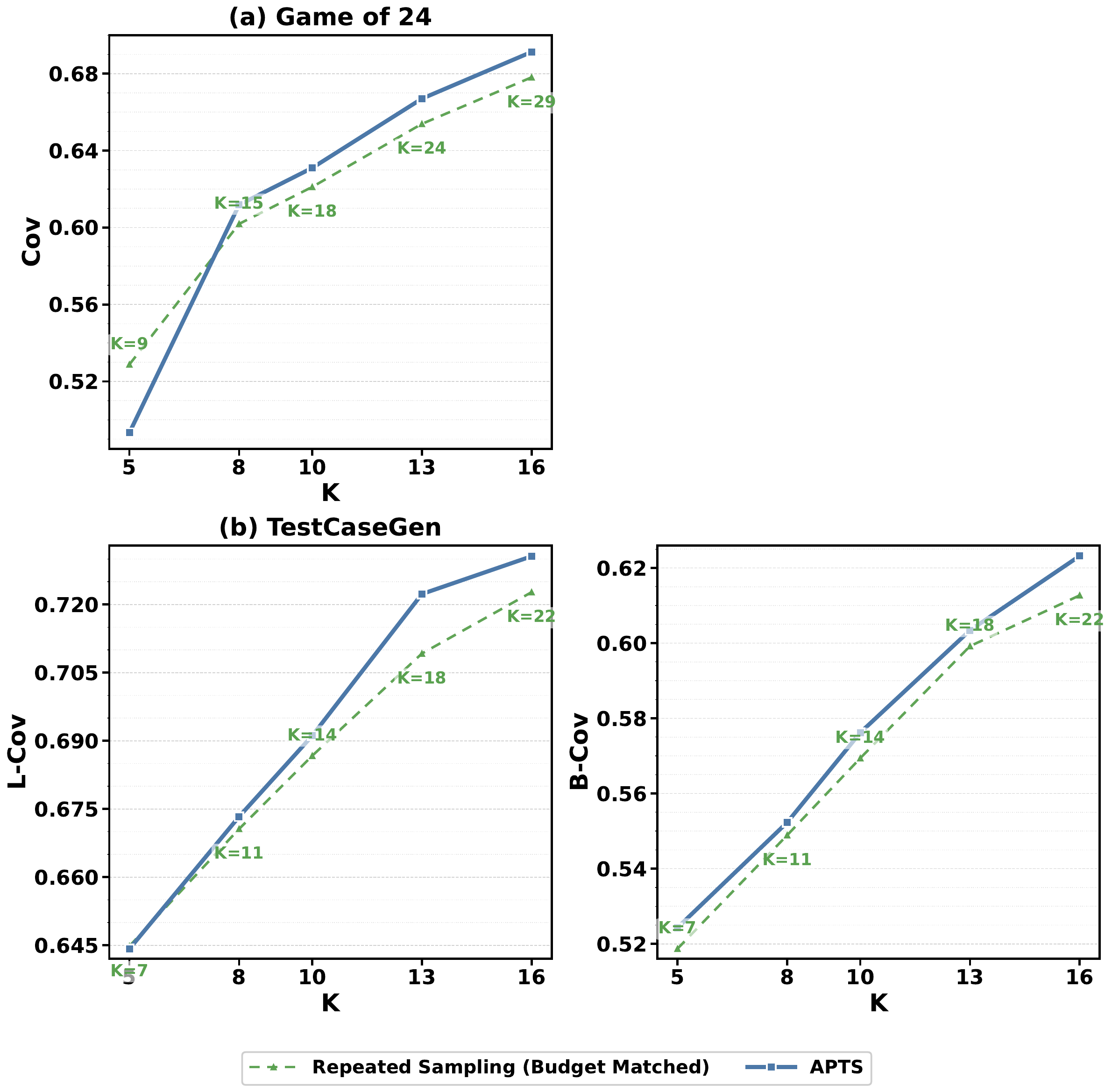}
    \caption{Budget-matched comparison on Qwen3-8B.}
    \label{fig:Samebudget}
\end{figure}

\paragraph{Budget-Matched Comparison.}
\label{sec:budget_matched_comparison}
Since APTS incurs additional token overhead due to tree search, a natural question is whether its gains could be achieved simply by spending the same budget on more repeated samples. To answer this, we conduct a budget-matched comparison by increasing the number of samples in Repeated Sampling until its total token cost matches that of APTS. 
Figure~\ref{fig:Samebudget} presents the results on Game of 24 and TestCaseGen. We exclude Forward Synthesis from this analysis because its evaluation metric, N-Circle, is strongly correlated with the number of sampled solutions: as $K$ increases, N-Circle naturally increases as well, making it difficult to disentangle improvements due to more effective search from those caused simply by returning more samples.

When $K$ is small (\eg, $K=5$), budget-matched Repeated Sampling can be comparable to or even slightly better than APTS, likely because direct sampling is still generate diverse solutions at low sample counts. However, as $K$ grows, the marginal benefit of repeated sampling decreases due to increasingly frequent duplicate solutions, whereas APTS can make better use of the same budget by explicitly exploring alternative reasoning trajectories. As a result, APTS consistently outperforms budget-matched Repeated Sampling at larger $K$ on both tasks. These results suggest that the gains of APTS do not arise merely from using more decoding tokens, but from answer-probing-guided search. 

More importantly, APTS is more sample-efficient. For example, on Game of 24, APTS achieves higher solution coverage using only 16 final answers than budget-matched Repeated Sampling using 29 samples (0.6913 vs.\ 0.6781). Similarly, on TestCaseGen, APTS achieves higher line coverage and branch coverage with 16 generated test cases than budget-matched Repeated Sampling with 22 samples (0.7306 vs.\ 0.7227 in L-Cov and 0.6232 vs.\ 0.6127 in B-Cov). This is particularly desirable in applications where each generated candidate must be further validated, executed, or reranked.

\section{Case Study and Error Analysis}
\label{sec:case_study}

We present two representative failure cases from Game of 24 on Qwen3-8B ($\alpha=0.7$) to provide deeper insights into APTS's behavior.

\paragraph{Case 1: Insufficient exploration under current $\alpha$.}
For the problem [2, 6, 10, 12], there are 9 solution categories. As shown in Table~\ref{tab:case_alpha}, APTS covers only 3 of them, with the remaining 6 categories unexplored. The 16 answers concentrate on a few dominant categories (Cat 2 appears 7 times), while categories involving more creative use of division (\eg $(10+2)\times12/6$, $(12/6+10)\times2$) are not discovered. This suggests that the current $\alpha$ may not provide a sufficiently strong diversity incentive for problems with many solution categories, and increasing $\alpha$ could encourage broader exploration.

\begin{table}[htb]
\centering
\small
\caption{APTS outputs for problem [2, 6, 10, 12] (9 categories, 3 covered, $\alpha\!=\!0.7$).}
\label{tab:case_alpha}
\begin{tabular}{cl}
\toprule
\textbf{Index} & \textbf{APTS Output} \\
\midrule
0 & $(12-10)\times6\times2$ \\
1 & $(10-6)\times12/2$ \\
2--5 & $(12-10)\times6\times2$ \\
6 & $(12+2-10)\times6$ \\
7--8 & $(10-6)\times12/2$ \\
9 & Error \\
10 & $(12+2-10)\times6$ \\
11 & Error \\
12 & $(10-6)\times12/2$ \\
13--15 & $(12-10)\times6\times2$ \\
\bottomrule
\end{tabular}
\end{table}

\paragraph{Case 2: Limited solution diversity in sampling.}
For the problem [2, 2, 4, 6], there are 5 solution categories, including $(4+2)\times(6-2)$, $(6+4+2)\times2$, and $(4-2)\times6\times2$. However, as shown in Table~\ref{tab:case_sampling}, all 16 APTS answers produce the identical expression $6\times4+2-2$, which exploits a trivial cancellation pattern ($a\times b+c-c$). This pattern recurs across problems with repeated numbers (\eg [2,6,6,12] $\rightarrow$ $12\times2+6-6$; [4,6,8,8] $\rightarrow$ $6\times4+8-8$). In these cases, the trivial cancellation acts as a strong attractor in the model's sampling behavior, and APTS cannot steer the model toward alternative solutions regardless of $\alpha$.

\begin{table}[htb]
\centering
\small
\caption{APTS outputs for problem [2, 2, 4, 6] (5 categories, 1 covered, $\alpha\!=\!0.7$).}
\label{tab:case_sampling}
\begin{tabular}{cl}
\toprule
\textbf{Index} & \textbf{APTS Output} \\
\midrule
0--15 & $6\times4+2-2$ \\
\bottomrule
\end{tabular}
\end{table}


\newpage
\section{Additional Results}

\begin{figure}[tbh]
    \centering
    \includegraphics[width=\linewidth]{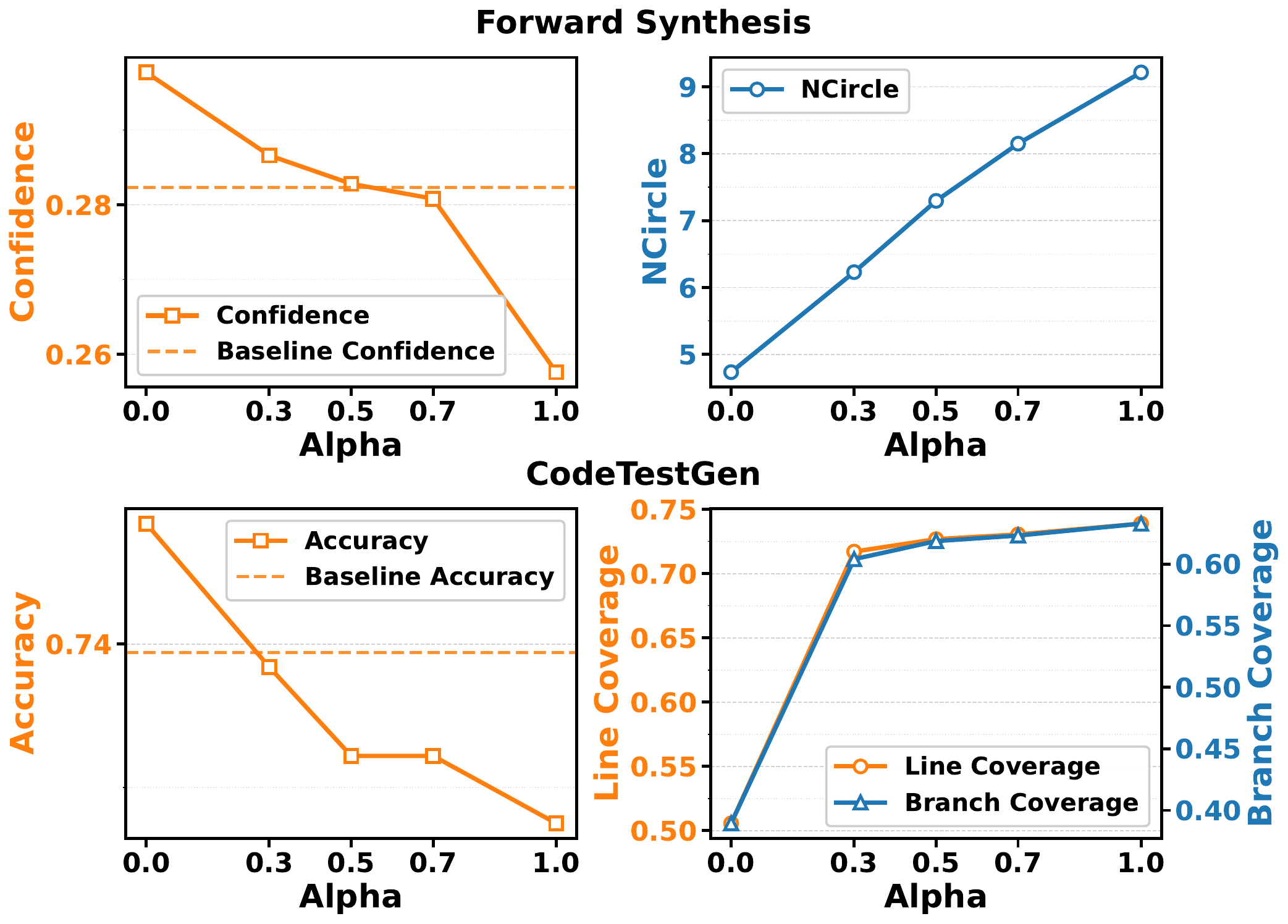}
    \caption{Effect of the trade-off coefficient $\alpha$ on Forward Synthesis and TestCaseGen with Qwen3-8B.}
    \label{fig:alpha_appendix}
\end{figure}

\begin{figure}[tbh]
    \centering
    \includegraphics[width=\linewidth]{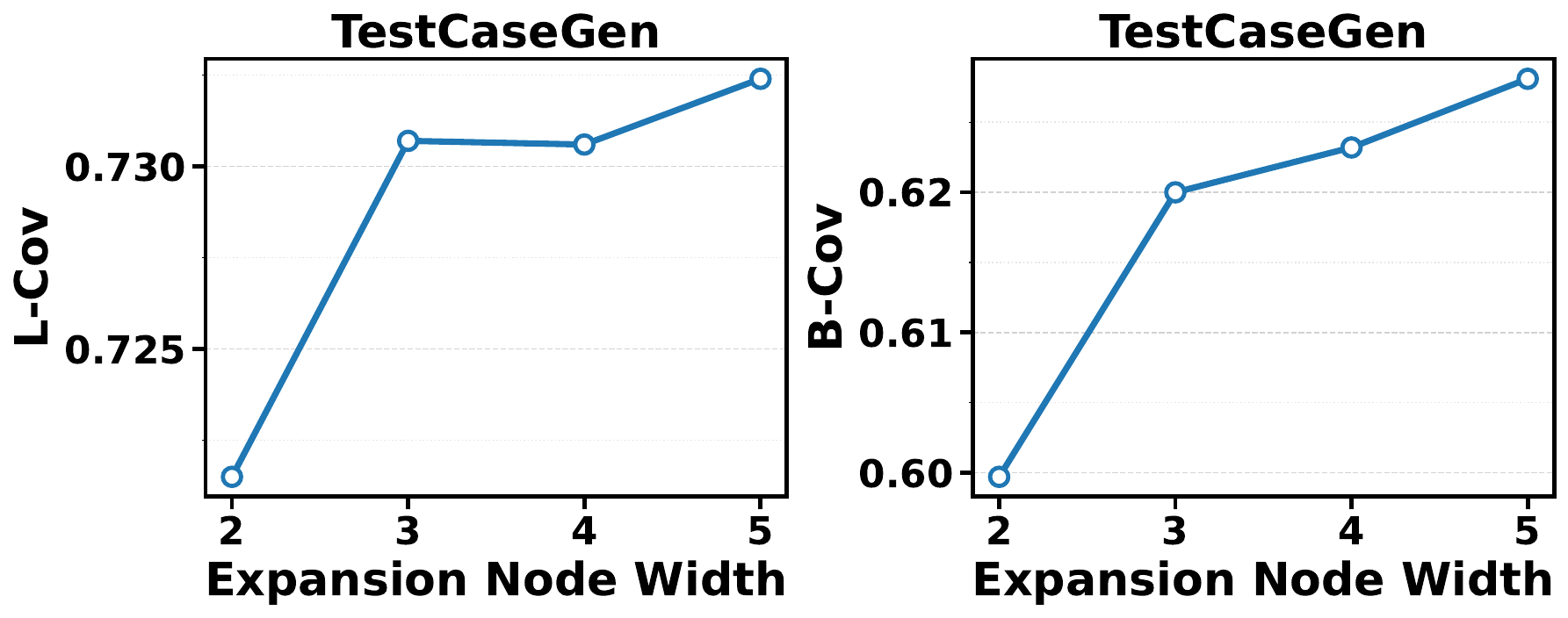}
    \caption{Effect of node expansion width on the solution
diversity of APTS on TestCaseGen with Qwen3-8B}
    \label{fig:nbw_code}
\end{figure}

\begin{figure}[tbh]
    \centering
    \includegraphics[width=\linewidth]{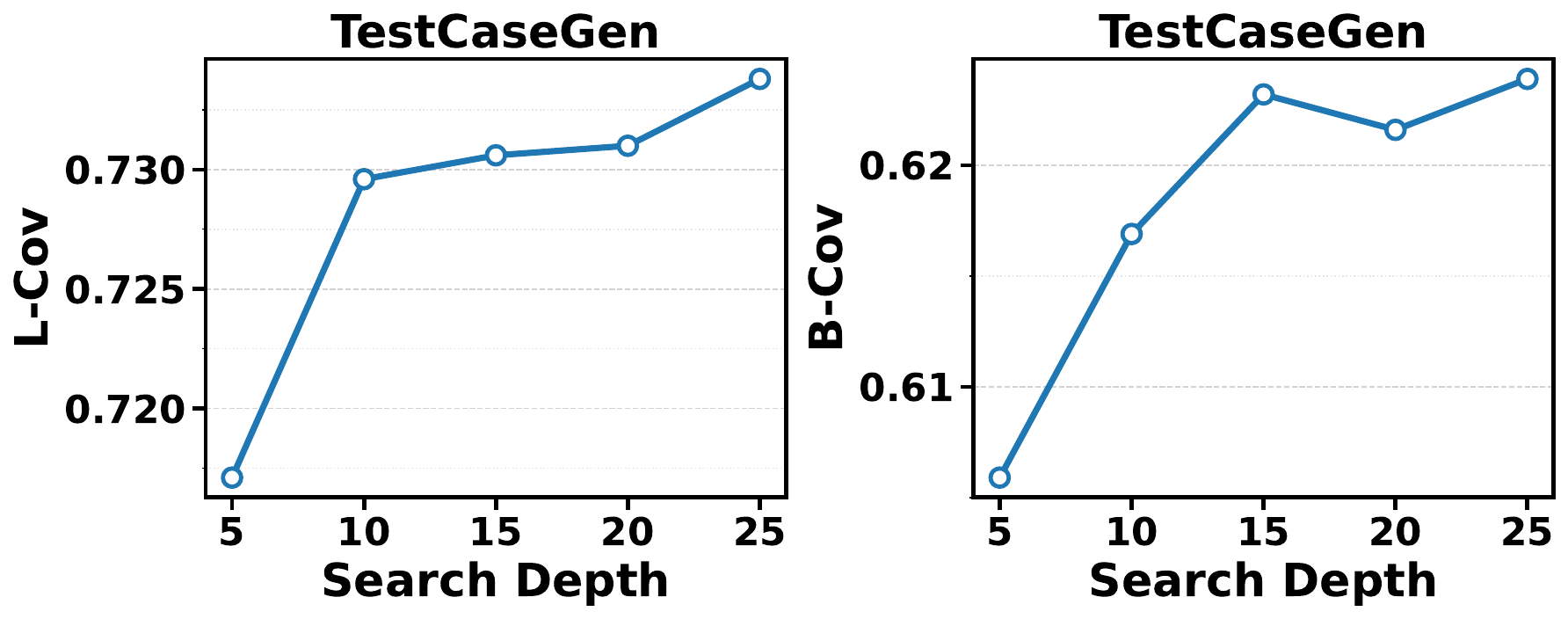}
    \caption{Effect of max search depth on the solution
diversity of APTS on TestCaseGen with Qwen3-8B}
    \label{fig:nbw_depth}
\end{figure}

\begin{figure*}[t]
    \centering
    \begin{subfigure}{\linewidth}
        \centering
        \caption{Problem (1, 3, 4, 8)}
        \includegraphics[width=\linewidth]{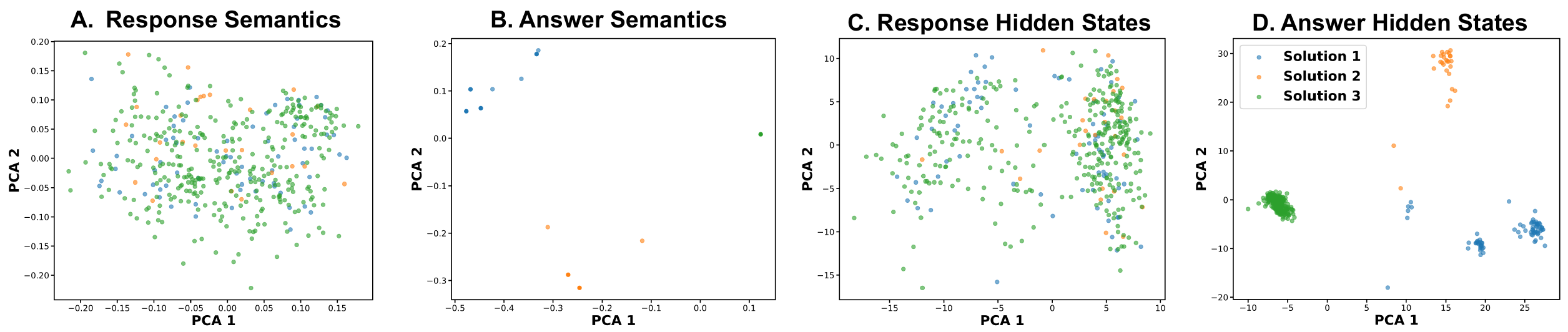}
        \label{fig:clustering_1348}
    \end{subfigure}

    \vspace{0.5em}

    \begin{subfigure}{\linewidth}
        \centering
        \caption{Problem (2, 6, 7, 9)}
        \includegraphics[width=\linewidth]{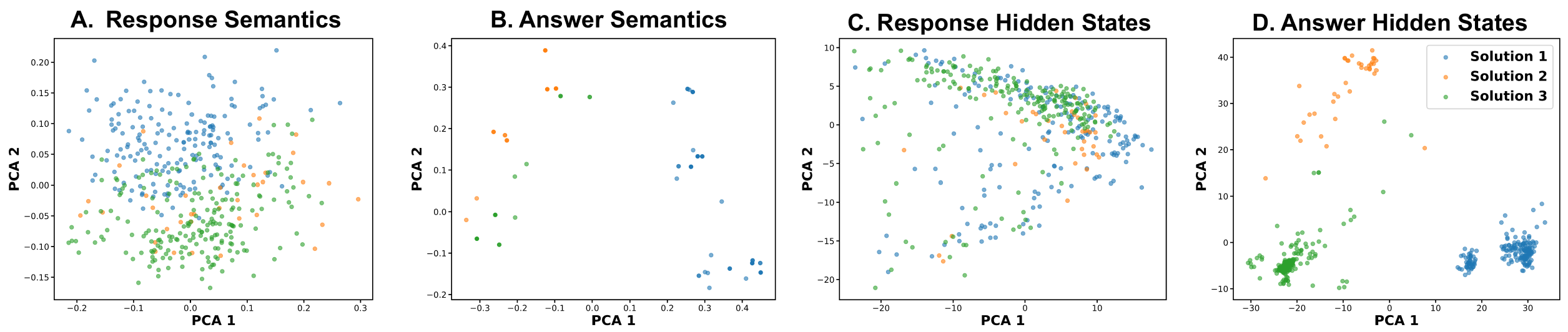}
        \label{fig:clustering_2679}
    \end{subfigure}

    \vspace{0.5em}

    \begin{subfigure}{\linewidth}
        \centering
        \caption{Problem (5, 5, 7, 7)}
        \includegraphics[width=\linewidth]{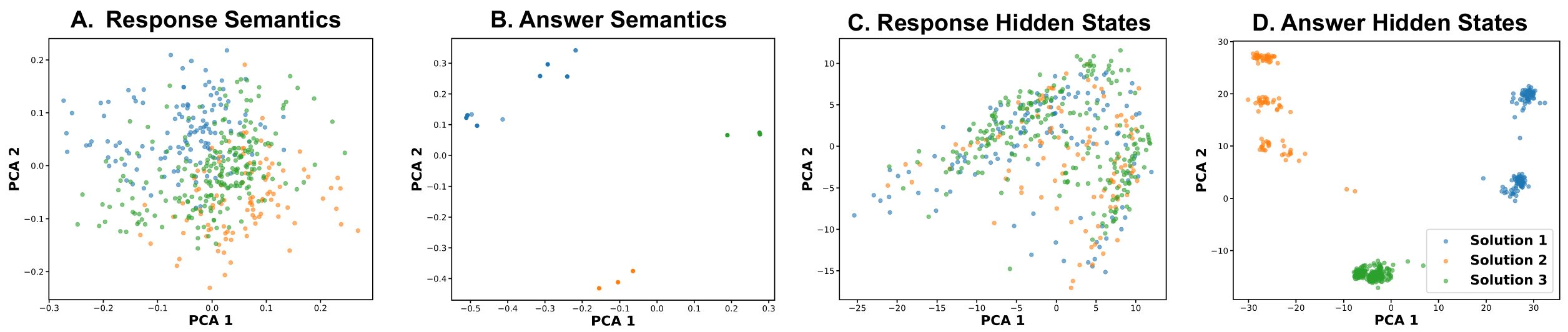}
        \label{fig:clustering_5577}
    \end{subfigure}

    \caption{PCA visualizations of solution-path clustering across different representation types.}
    \label{fig:clustering_all}
\end{figure*}

\begin{figure*}[t]
    \centering
    \begin{subfigure}[t]{0.32\linewidth}
        \centering
        \caption{Problem (1, 3, 4, 8)}
        \includegraphics[width=\linewidth]{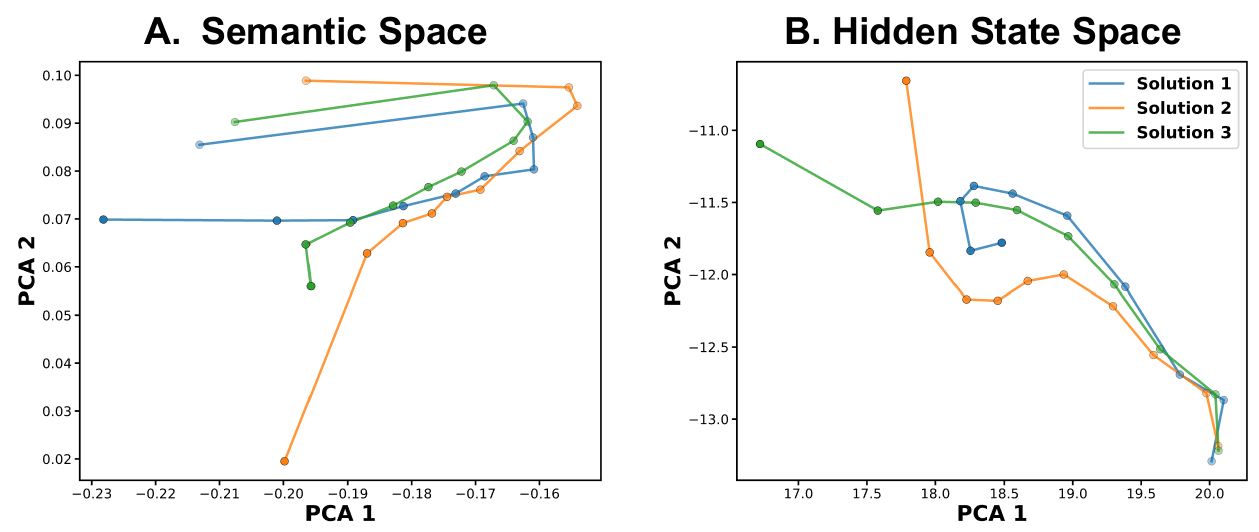}
        \label{fig:trajectory_1348}
    \end{subfigure}
    \hfill
    \begin{subfigure}[t]{0.32\linewidth}
        \centering
        \caption{Problem (2, 6, 7, 9)}
        \includegraphics[width=\linewidth]{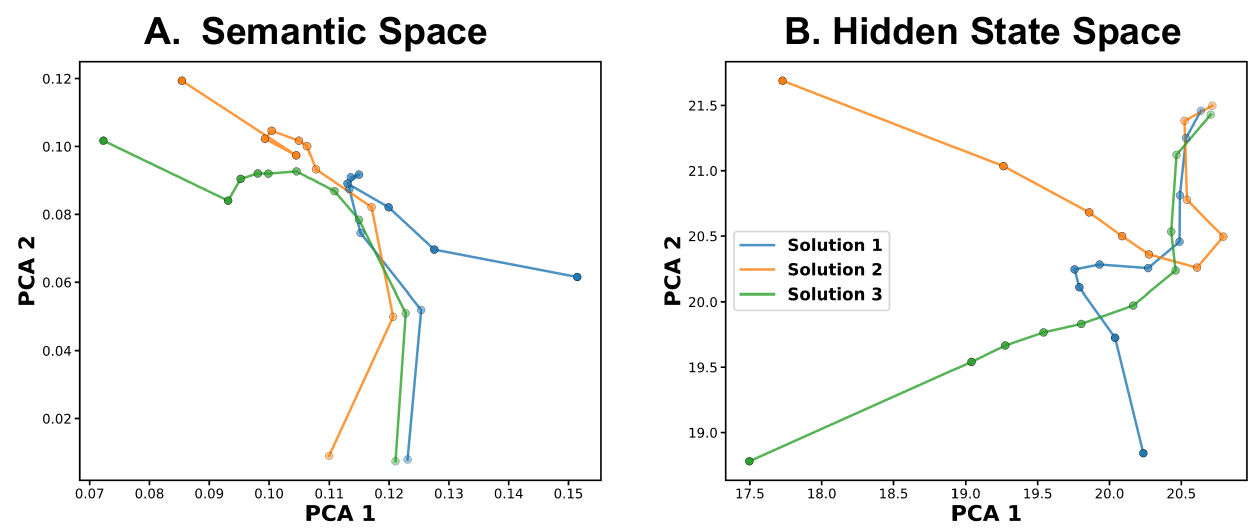}
        \label{fig:trajectory_2679}
    \end{subfigure}
    \hfill
    \begin{subfigure}[t]{0.32\linewidth}
        \centering
        \caption{Problem (5, 5, 7, 7)}
        \includegraphics[width=\linewidth]{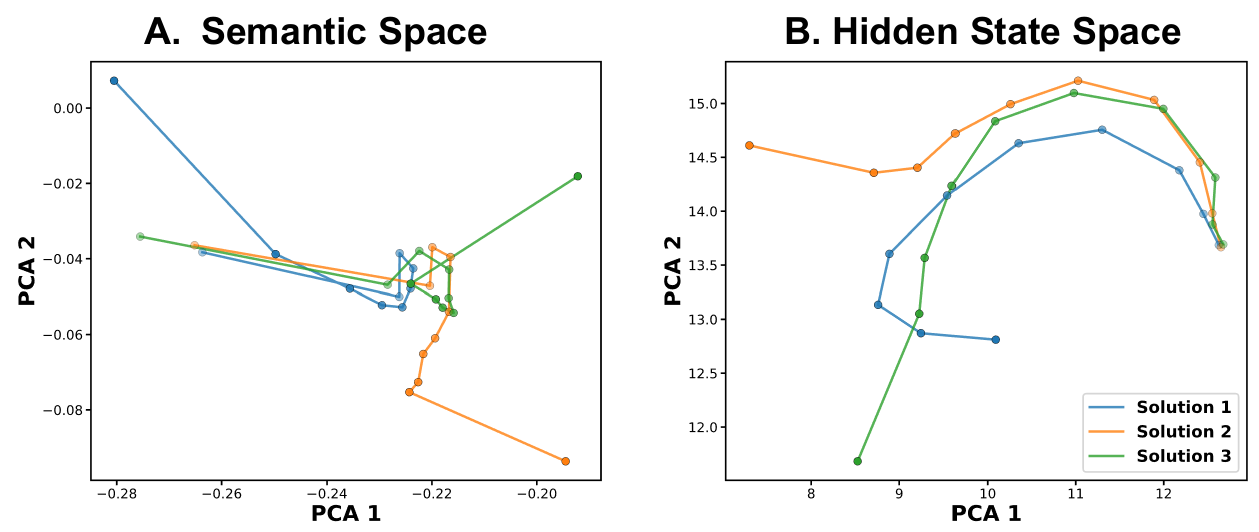}
        \label{fig:trajectory_5577}
    \end{subfigure}
    \caption{Visualizations of solution-path trajectories in semantic and hidden-state spaces for different problem groups.}
    \label{fig:trajectory_all}
\end{figure*}

\begin{algorithm*}[t]
\caption{APTS Algorithm}
\label{alg:apts}
\renewcommand{\algorithmicrequire}{\textbf{Input:}}
\renewcommand{\algorithmicensure}{\textbf{Output:}}
\begin{algorithmic}[1]
\REQUIRE Problem $p$, target response number $M$, node expansion width $W$, maximum search depth $D$, trade-off coefficient $\alpha$

\STATE $\mathcal{N}^{(0)} \leftarrow \operatorname{LLM}(p, M) = \{n_1^{(0)}, \dots, n_M^{(0)}\}$
\FOR{$d = 0$ to $D-1$}
    \IF{all nodes in $\mathcal{N}^{(d)}$ are complete}
        \STATE \textbf{break}
    \ENDIF
    \STATE
    \STATE \text{\#\#\#\# Expansion Stage \#\#\#\#}
    \STATE $\mathcal{C}^{(d+1)} \leftarrow \emptyset$
    \FOR{each $n_i^{(d)} \in \mathcal{N}^{(d)}$}
        \STATE $\text{Children}\!\left(n_i^{(d)}\right) \leftarrow \operatorname{GenerateNextStep}(p \oplus n_i^{(d)}, W)$
        \STATE $\mathcal{C}^{(d+1)} \leftarrow \mathcal{C}^{(d+1)} \cup \text{Children}\!\left(n_i^{(d)}\right)$
    \ENDFOR

    \FOR{each $c_i \in \mathcal{C}^{(d+1)}$}
        \STATE $(H_i, \mathrm{PPL}_i) \leftarrow \textbf{AnswerProbing}(p, c_i)$
    \ENDFOR

    \STATE
    \STATE \text{\#\#\#\# Selection Stage \#\#\#\#}
    \STATE $\mathcal{S} \leftarrow \emptyset$
    \FOR{$m = 1$ to $M$}
        \FOR{each $c_i \in \mathcal{C}^{(d+1)} \setminus \mathcal{S}$}
            \IF{$\mathcal{S} = \emptyset$}
                \STATE $DS_i \leftarrow 0$
            \ELSE
                \STATE $DS_i \leftarrow -\max_{c_j \in \mathcal{S}} \cos(H_i, H_j)$
            \ENDIF
            \STATE $V(c_i) \leftarrow (1-\alpha) / \mathrm{PPL}_i + \alpha DS_i$
        \ENDFOR
        \STATE $\mathcal{S} \leftarrow \mathcal{S} \cup \{\arg\max_{c_i \in \mathcal{C}^{(d+1)} \setminus \mathcal{S}} V(c_i)\}$
    \ENDFOR

    \STATE $\mathcal{N}^{(d+1)} \leftarrow \mathcal{S}$
\ENDFOR
\FOR{each $n_i^{(d)} \in \mathcal{N}^{(d)}$}
    \STATE Complete the response from $p \oplus n_i^{(d)}$
\ENDFOR
\ENSURE the resulting $M$ responses
\end{algorithmic}
\end{algorithm*}

%% file: main.bib
@inproceedings{JainSR25,
  author       = {Kush Jain and
                  Gabriel Synnaeve and
                  Baptiste Rozi{\`{e}}re},
  title        = {TestGenEval: {A} Real World Unit Test Generation and Test Completion
                  Benchmark},
  booktitle    = {{ICLR}},
  year         = {2025}
}

@article{drug_discover,
author = {Jürgen Drews },
title = {Drug Discovery: A Historical Perspective},
journal = {Science},
year = {2000},
}

@article{RL_Sim,
  author       = {Yuda Song and
                  Julia Kempe and
                  R{\'{e}}mi Munos},
  title        = {Outcome-based Exploration for {LLM} Reasoning},
  journal      = {CoRR},
  volume       = {abs/2509.06941},
  year         = {2025}
}

@article{jiang2025artificial,
  title={Artificial hivemind: The open-ended homogeneity of language models (and beyond)},
  author={Jiang, Liwei and Chai, Yuanjun and Li, Margaret and Liu, Mickel and Fok, Raymond and Dziri, Nouha and Tsvetkov, Yulia and Sap, Maarten and Albalak, Alon and Choi, Yejin},
  journal={NIPS},
  year={2025}
}

@inproceedings{renze2024effect,
  title={The effect of sampling temperature on problem solving in large language models},
  author={Renze, Matthew},
  booktitle={{EMNLP} (Findings)},
  year={2024}
}

@inproceedings{HoltzmanBDFC20,
  author       = {Ari Holtzman and
                  Jan Buys and
                  Li Du and
                  Maxwell Forbes and
                  Yejin Choi},
  title        = {The Curious Case of Neural Text Degeneration},
  booktitle    = {{ICLR}},
  year         = {2020}
}

@inproceedings{YunAWPS25,
  author       = {Longfei Yun and
                  Chenyang An and
                  Zilong Wang and
                  Letian Peng and
                  Jingbo Shang},
  title        = {The Price of Format: Diversity Collapse in LLMs},
  booktitle    = {{EMNLP} (Findings)},
  year         = {2025}
}

@misc{DiverseBeam,
      title={Diverse Beam Search: Decoding Diverse Solutions from Neural Sequence Models}, 
      author={Ashwin K Vijayakumar and Michael Cogswell and Ramprasath R. Selvaraju and Qing Sun and Stefan Lee and David Crandall and Dhruv Batra},
      year={2018},
      booktitle= {AAAI},
}

@misc{SemDiD,
      title={Semantic-guided Diverse Decoding for Large Language Model}, 
      author={Weijie Shi and Yue Cui and Yaguang Wu and Jingzhi Fang and Shibo Zhang and Mengze Li and Sirui Han and Jia Zhu and Jiajie Xu and Xiaofang Zhou},
      booktitle    = {NIPS},
      year         = {2025}
}

@article{ToT,
  title={Tree of thoughts: Deliberate problem solving with large language models},
  author={Yao, Shunyu and Yu, Dian and Zhao, Jeffrey and Shafran, Izhak and Griffiths, Tom and Cao, Yuan and Narasimhan, Karthik},
  journal={NIPS},
  year={2023}
}

@article{beam_search,
  title={A simple, fast diverse decoding algorithm for neural generation},
  author={Li, Jiwei and Monroe, Will and Jurafsky, Dan},
  journal={arXiv preprint arXiv:1611.08562},
  year={2016}
}

@article{kirk2023understanding,
  title={Understanding the effects of rlhf on llm generalisation and diversity},
  author={Kirk, Robert and Mediratta, Ishita and Nalmpantis, Christoforos and Luketina, Jelena and Hambro, Eric and Grefenstette, Edward and Raileanu, Roberta},
  journal={ICLR},
  year={2024}
}

@inproceedings{o2024attributing,
  title={Attributing mode collapse in the fine-tuning of large language models},
  author={O’Mahony, Laura and Grinsztajn, Leo and Schoelkopf, Hailey and Biderman, Stella},
  booktitle={ICLR 2024 Workshop on Mathematical and Empirical Understanding of Foundation Models},
  year={2024}
}

@misc{yang2025qwen3technicalreport,
      title={Qwen3 Technical Report}, 
      author={Qwen Team},
      journal={arXiv preprint arXiv:2505.09388},
      year={2025},
}

@article{qwen3embedding,
  title={Qwen3 Embedding: Advancing Text Embedding and Reranking Through Foundation Models},
  author={Zhang, Yanzhao and Li, Mingxin and Long, Dingkun and Zhang, Xin and Lin, Huan and Yang, Baosong and Xie, Pengjun and Yang, An and Liu, Dayiheng and Lin, Junyang and Huang, Fei and Zhou, Jingren},
  journal={arXiv preprint arXiv:2506.05176},
  year={2025}
}

@misc{fang2026controllablellmreasoningsparse,
      title={Controllable LLM Reasoning via Sparse Autoencoder-Based Steering}, 
      author={Yi Fang and Wenjie Wang and Mingfeng Xue and Boyi Deng and Fengli Xu and Dayiheng Liu and Fuli Feng},
      journal={arXiv preprint arXiv:2601.03595},
      year={2026},
}

@article{reasong_strength,
  author       = {Leheng Sheng and
                  An Zhang and
                  Zijian Wu and
                  Weixiang Zhao and
                  Changshuo Shen and
                  Yi Zhang and
                  Xiang Wang and
                  Tat{-}Seng Chua},
  title        = {On Reasoning Strength Planning in Large Reasoning Models},
  journal    = {NeurIPS},
  year         = {2025}
}

@article{huang2025benchmarking,
  title={Benchmarking llms for unit test generation from real-world functions},
  author={Huang, Dong and Zhang, Jie M and Harman, Mark and Zhang, Qianru and Du, Mingzhe and Ng, See-Kiong},
  journal={arXiv preprint arXiv:2508.00408},
  year={2025}
}

@article{yu2024llasmol,
  title={Llasmol: Advancing large language models for chemistry with a large-scale, comprehensive, high-quality instruction tuning dataset},
  author={Yu, Botao and Baker, Frazier N and Chen, Ziqi and Ning, Xia and Sun, Huan},
  journal={COLM},
  year={2024}
}

@misc{openai2025gptoss120bgptoss20bmodel,
      title={gpt-oss-120b \& gpt-oss-20b Model Card}, 
      author={OpenAI},
      journal={arXiv preprint arXiv:2508.10925},
      year={2025}
}

@article{wang2025testeval,
  title={Testeval: Benchmarking large language models for test case generation},
  author={Wang, Wenhan and Yang, Chenyuan and Wang, Zhijie and Huang, Yuheng and Chu, Zhaoyang and Song, Da and Zhang, Lingming and Chen, An Ran and Ma, Lei},
  journal={{NAACL} (Findings)},
  year={2025}
}

@article{xie2021much,
  title={How much space has been explored? measuring the chemical space covered by databases and machine-generated molecules},
  author={Xie, Yutong and Xu, Ziqiao and Ma, Jiaqi and Mei, Qiaozhu},
  journal={ICLR},
  year={2023}
}

@article{bajusz2015tanimoto,
  title={Why is Tanimoto index an appropriate choice for fingerprint-based similarity calculations?},
  author={Bajusz, D{\'a}vid and R{\'a}cz, Anita and H{\'e}berger, K{\'a}roly},
  journal={Journal of cheminformatics},
  volume={7},
  number={1},
  pages={20},
  year={2015},
  publisher={Springer}
}

@article{schwaller2021extraction,
  title={Extraction of organic chemistry grammar from unsupervised learning of chemical reactions},
  author={Schwaller, Philippe and Hoover, Benjamin and Reymond, Jean-Louis and Strobelt, Hendrik and Laino, Teodoro},
  journal={Science Advances},
  year={2021},
}

@article{prompt_diverse_1,
  author       = {Ranjita Naik and
                  Varun Chandrasekaran and
                  Mert Y{\"{u}}ksekg{\"{o}}n{\"{u}}l and
                  Hamid Palangi and
                  Besmira Nushi},
  title        = {Diversity of Thought Improves Reasoning Abilities of Large Language
                  Models},
  journal      = {arXiv preprint arXiv:2310.07088},
  year         = {2023}
}

@article{ruan2026evaluating,
  title     = {Evaluating LLMs' divergent thinking capabilities for scientific idea generation with minimal context},
  author    = {Ruan, Kai and Wang, Xuan and Hong, Jixiang and Wang, Peng and Liu, Yang and Sun, Hao},
  journal   = {Nature communications},
  year      = {2026},
}

@article{farquhar2024detecting,
  title     = {Detecting hallucinations in large language models using semantic entropy},
  author    = {Farquhar, Sebastian and Kossen, Jannik and Kuhn, Lorenz and Gal, Yarin},
  journal   = {Nature},
  year      = {2024},
}

@inproceedings{Certaindex,
  author    = {Fu, Yichao and Chen, Junda and Zhu, Siqi and Fu, Fu and Dai, Zhongdongming and Zhuang, Yonghao and Ma, Yian and Qiao, Aurick and Rosing, Tajana S and Stoica, Ion and Zhang, Hao},
  booktitle = {NeurIPS},
  title     = {Efficiently Scaling LLM Reasoning Programs with Certaindex},
  year      = {2025}
}

@inproceedings{SDE2,
  title     = {{SD}-E2: Semantic Exploration for Reasoning Under Token Budgets},
  author    = {Mishra, Kshitij  and
               Lukas, Nils  and
               Lahlou, Salem},
  booktitle = {Findings of the {A}ssociation for {C}omputational {L}inguistics: {EACL} 2026},
  year      = {2026},
}

@misc{VERL,
  title         = {Semantic-Space Exploration and Exploitation in RLVR for LLM Reasoning},
  author        = {Fanding Huang and Guanbo Huang and Xiao Fan and Yi He and Xiao Liang and Xiao Chen and Qinting Jiang and Faisal Nadeem Khan and Jingyan Jiang and Zhi Wang},
  year          = {2026},
  journal={arXiv preprint arXiv:2509.23808},
}

@inproceedings{TreeGRPO,
  title     = {Tree{GRPO}: Tree-Advantage {GRPO} for Online {RL} Post-Training of Diffusion Models},
  author    = {Zheng Ding and Weirui Ye},
  booktitle = {ICLR},
  year      = {2026},
}

@inproceedings{SPOtree,
  title     = {Segment Policy Optimization: Effective Segment-Level Credit Assignment in {RL} for Large Language Models},
  author    = {Yiran Guo and Lijie Xu and Jie Liu and Ye Dan and Shuang Qiu},
  booktitle = {NeurIPS},
  year      = {2025},
}

@inproceedings{GuidedSampling,
  title     = {GuidedSampling: Steering {LLM}s Towards Diverse Candidate Solutions at Inference-Time},
  author    = {Divij Handa and Mihir Parmar and Aswin RRV and Md Nayem Uddin and Hamid Palangi and Chitta Baral},
  booktitle = {ICLR},
  year      = {2026},
}

@inproceedings{STARS,
  title     = {Exploring Diverse Generation Paths via Inference-time Stiefel Activation Steering},
  author    = {Dongxuan Zhu and Ly Tran Ho Khanh and Andy Yat-Ming Cheung and Man-Chung Yue and Viet Anh Nguyen},
  booktitle = {ICLR},
  year      = {2026},
}
